\documentclass[10pt,twocolumn,letterpaper]{article}

\usepackage{iccv}
\usepackage{times}
\usepackage{epsfig}

\usepackage{graphicx}
\usepackage{amsmath}
\usepackage{amssymb}
\usepackage{booktabs}

\usepackage{algorithm}
\usepackage{algorithmic}
\usepackage{listings} 
 
\usepackage{subcaption}
\usepackage{bm}
\usepackage{bbm}
\usepackage{multicol}
\usepackage{multirow}
\usepackage{color}
\usepackage{caption}
\usepackage{hhline}
\usepackage{pifont}
\usepackage{threeparttable}
\usepackage{makecell}
\usepackage{wrapfig}
\usepackage{colortbl}  
\usepackage{url}
\usepackage{enumitem}
\usepackage{tablefootnote}
\usepackage[table]{xcolor}
\usepackage[title]{appendix}

\definecolor{neucolor}{RGB}{160,160,160}
\definecolor{bblue}{rgb}{0.855,0.933,0.98}

\newcommand{\graybox}[1]{\colorbox{lightgray!28}{#1}}

\newcommand{\tabbl}{\midrule \rowcolor{lightgray!17}}

\newcommand{\xmark}{\ding{55}}
\newcommand{\cmark}{\ding{51}}
\newcommand{\myparagraph}[1]{{\vspace{.4em} \noindent \bf #1}}

\definecolor{codeblue}{rgb}{0.25,0.5,0.5}
\definecolor{codegreen}{rgb}{0,0.6,0}
\definecolor{codekw}{rgb}{0.85, 0.18, 0.50}
\usepackage{etoolbox}
\makeatletter
\AfterEndEnvironment{algorithm}{\let\@algcomment\relax}
\AtEndEnvironment{algorithm}{\kern2pt\hrule\relax\vskip3pt\@algcomment}
\let\@algcomment\relax
\newcommand\algcomment[1]{\def\@algcomment{\footnotesize#1}}
\renewcommand\fs@ruled{\def\@fs@cfont{\bfseries}\let\@fs@capt\floatc@ruled
  \def\@fs@pre{\hrule height.8pt depth0pt \kern2pt}%
  \def\@fs@post{}%
  \def\@fs@mid{\kern2pt\hrule\kern2pt}%
  \let\@fs@iftopcapt\iftrue}
\makeatother

\definecolor{redstar}{RGB}{240,47,29}
\definecolor{curly}{RGB}{187,101,183}
\definecolor{lightsalmon}{RGB}{255,160,122}
\definecolor{limegreen}{RGB}{50,205,50}
\definecolor{grey}{RGB}{190,190,190}
\definecolor{poscolor}{RGB}{83,161,81}
\definecolor{negcolor}{RGB}{103,81,165}
\definecolor{neucolor}{RGB}{190,190,190}
\newcommand{\posval}[1]{{\textbf{\footnotesize\selectfont \color{poscolor}~($+$#1)}}}
\newcommand{\negval}[1]{{\textbf{\footnotesize\selectfont \color{negcolor}~($-$#1)}}}

\newcommand{\modelname}{SWORD}

\usepackage[pagebackref=true,breaklinks=true,letterpaper=true,colorlinks=True,bookmarks=false]{hyperref}

\usepackage[capitalize]{cleveref}
\crefname{section}{Sec.}{Secs.}
\Crefname{section}{Section}{Sections}
\Crefname{table}{Table}{Tables}
\crefname{table}{Tab.}{Tabs.}

\iccvfinalcopy 

\def\iccvPaperID{xxxx} 

\begin{document}

\title{Exploring Transformers for Open-world Instance Segmentation}

\author
{
Jiannan Wu$^{1}$, 
~~~
Yi Jiang$^{2}$,
~~~
Bin Yan$^{3}$, 
~~~
Huchuan Lu$^{3}$, 
~~~
Zehuan Yuan$^{2}$, 
~~~
Ping Luo$^{1,4}$
\\[0.2cm]
${^1}$The University of Hong Kong ~~~
${^2}$ByteDance ~~~ \\[0.1cm]
${^3}$Dalian University of Technology ~~~
${^4}$Shanghai AI Laboratory 
}

\ificcvfinal\thispagestyle{empty}\fi

\twocolumn[{%
\renewcommand\twocolumn[1][]{#1}%
\maketitle
\begin{center}
    \centering
    \captionsetup{type=figure}
    \includegraphics[width=\textwidth]{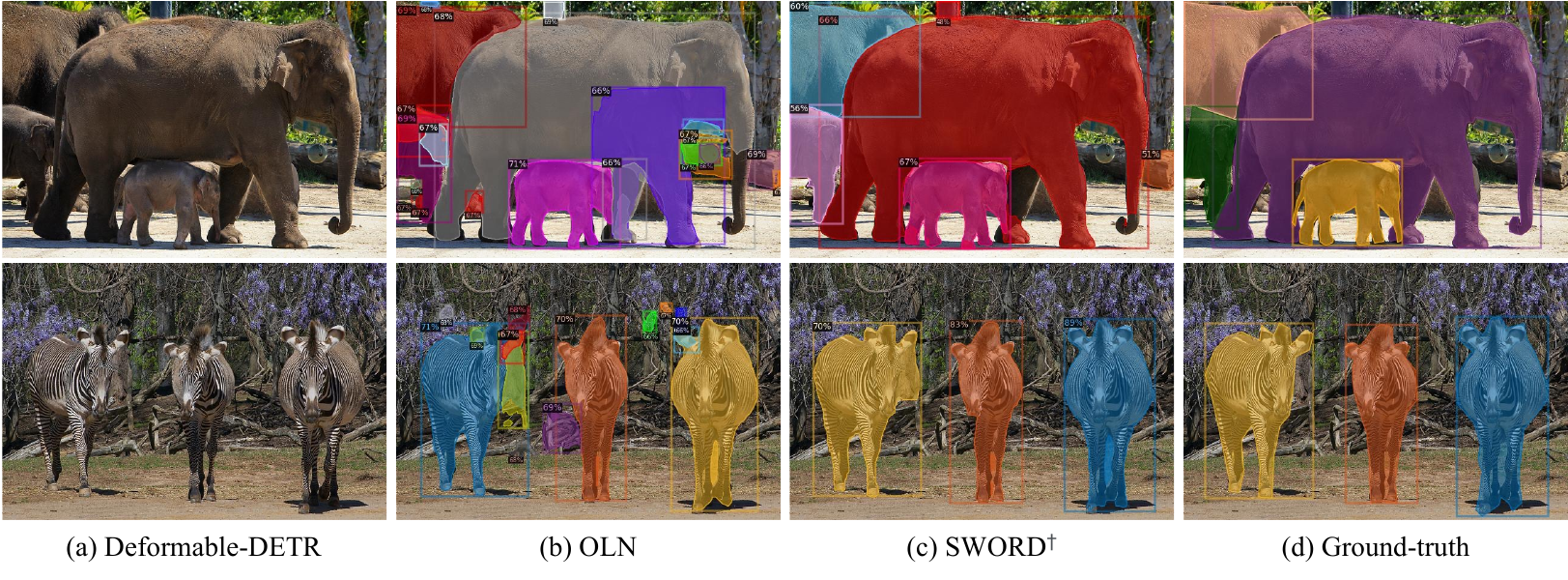}
    \vspace{-6mm}
    \caption{{\bf Comparison of \modelname{}$^{\dag}$ with different methods}. The images are from the validation set of COCO~\cite{lin2014coco}. All the models are trained on the PASCAL-VOC~\cite{everingham2010pascal} classes of COCO dataset, where \textit{elephants} and \textit{zebras} are not seen during training. (a) Deformable-DETR~\cite{zhu2020deformable-detr} fails to segment the objects not labeled in the training set. (b) OLN~\cite{kim2022oln} could localize the \textit{novel} objects, however, it also produce many false positives. (c) Our proposed \modelname{}$^{\dag}$ can predict the correct and accurate masks for unseen categories.}
    \label{fig:1}
\end{center}%
}]

\begin{abstract} 
   Open-world instance segmentation is a rising task, which aims to segment all objects in the image by learning from a limited number of base-category objects. This task is challenging, as the number of unseen categories could be hundreds of times larger than that of seen categories. 
   Recently, the DETR-like models have been extensively studied in the closed world while stay unexplored in the open world. In this paper, we utilize the Transformer for open-world instance segmentation and present \modelname{}. Firstly, we introduce to attach the stop-gradient operation before classification head and further add IoU heads for discovering novel objects. We demonstrate that a simple stop-gradient operation not only prevents the novel objects from being suppressed as background, but also allows the network to enjoy the merit of heuristic label assignment. Secondly, we propose a novel contrastive learning framework to enlarge the representations between objects and background. Specifically, we maintain a universal object queue to obtain the object center, and dynamically select positive and negative samples from the object queries for contrastive learning.
   While the previous works only focus on pursuing average recall and neglect average precision, we show the prominence of \modelname{} by giving consideration to both criteria. Our models achieve state-of-the-art performance in various open-world cross-category and cross-dataset generalizations. Particularly, in VOC to non-VOC setup, our method sets new state-of-the-art results of 40.0\% on AR$_{100}^{\rm b}$ and 34.9\% on AR$_{100}^{\rm m}$. For COCO to UVO generalization, \modelname{} significantly outperforms the previous best open-world model by 5.9\% on AP$^{\rm m}$ and 8.1\% on AR$_{100}^{\rm m}$. 
\end{abstract}

\vspace{-1mm}
\section{Introduction}\label{sec:introduction}
\vspace{-1mm}

\begin{table}[ht]
    \centering
    \renewcommand\arraystretch{1.2} 
    \setlength{\tabcolsep}{4.5pt}    
    \small
    \caption{The open-world generalization setups, which are established by the recent advanced approaches~\cite{kim2022oln, saito2021ldet, wang2022ggn}. The values in the bracket indicate the class numbers.
    }
    \vspace{-4mm}
    \begin{tabular}{c|c|c|c|c}
\Xhline{0.7pt}
\textbf{Dataset} & \textbf{Train} & \textbf{Evaluate} & \textbf{Image} & \textbf{Mask} \\

\tabbl
\multicolumn{5}{l}{\small{\textbf{\emph{Cross-cateory Generalization}}}}  \\

COCO & VOC(20) & non-VOC(60) & 95k & 493k \\
LVIS & COCO(80) & non-COCO(1123) & 100k & 455k \\

\tabbl
\multicolumn{5}{l}{\small{\textbf{\emph{Cross-dataset Generalization}}}}  \\

UVO & COCO(80) & non-COCO(-) & 118k & 860k \\
Objects365 & COCO(80) & non-COCO(285)  & 118k & 860k \\

\Xhline{0.7pt}
\end{tabular}

    \label{tab:dataset} 
\end{table}

The standard instance segmentation models~\cite{he2017maskrcnn, tian2020condinst, cheng2022mask2former} are developed to segment the objects from a predefined taxonomy, which is not often reflective of the diversity of object classes encountered in the real world. Recently, class-agnostic \textbf{open-world instance segmentation} introduced by the advanced approaches~\cite{kim2022oln, saito2021ldet, wang2022ggn} has gained increasing attention in the community. It requires the models to segment all objects of arbitrary categories in the image while only base-category objects can be seen during training. This task is highly challenging, as the number of unseen categories can be orders of magnitude larger than the number of seen categories. As shown in the second row of Table~\ref{tab:dataset}, for COCO to LVIS setup, there are 1123 non-COCO classes for out-of-domain evaluation while only objects belonging to 80 COCO classes are annotated in the training set. Besides, a critical challenge in the open-world scenario is that the \textit{novel} objects and background co-exist in the un-annotated regions. Consequently, the closed-world instance segmentation models fail to recognize unseen objects (Figure~\ref{fig:1}\textcolor{red}{a}) as they equally treat the \textit{novel} objects and background as negative samples during training.

Recently, DETR-like~\cite{carion2020detr, zhu2020deformable-detr} models based on Transformers~\cite{vaswani2017transformer} have exhibited superior performance in standard object detection and instance segmentation tasks.  However, the study of these Transformer-based models in the field of open-world instance segmentation is still a blank page to the community, as the previous works have exclusively relied on the Mask-RCNN~\cite{he2017maskrcnn} architecture. In this work, we aim to fill in the gap by delving deeply into the recent advanced Deformable-DETR~\cite{zhu2020deformable-detr}. 

An inspiring open-world method OLN~\cite{kim2022oln} proposes a classification-free network and estimates the scores of regions purely by localization quality (\emph{e.g.}, IoU score). In this manner, \textit{novel} objects would not be penalized as background due to the absence of classification learning. And the localization quality score is proven to be a better objectness cue for discovering \textit{novel} objects. Inspired by this spirit, a straightforward solution for Transformer-based models in open-world instance segmentation is to replace the classification head with IoU heads. However, this could lead to \textbf{two negative effects}: (i) The Transformer-based models are optimized with set prediction loss~\cite{carion2020detr}, where the classification score is indispensible for label assignment. Therefore, simply removing the classification head could be harmful to the process. (ii) The network would inherit the limitation of OLN that lacks the discriminative ability to differentiate the objects and background. This is because OLN is only trained with positive samples and thus fails to perceive the background. As a result, it would produce numerous false positives (Figure~\ref{fig:1}\textcolor{red}{b}) and result in fairly low average precision (AP). For example, in COCO to UVO (all) generalization, AR$^{\rm m}_{100}$ of OLN~\cite{kim2022oln} increases from 36.7\% to 42.1\% while AP$^{\rm m}$ significantly drops from 20.7\% to 14.0\% when compared with Mask-RCNN~\cite{he2017maskrcnn}.

\begin{figure}[t]
\vspace{-2mm}
\hspace{-2mm}
\includegraphics[width=0.50\textwidth]{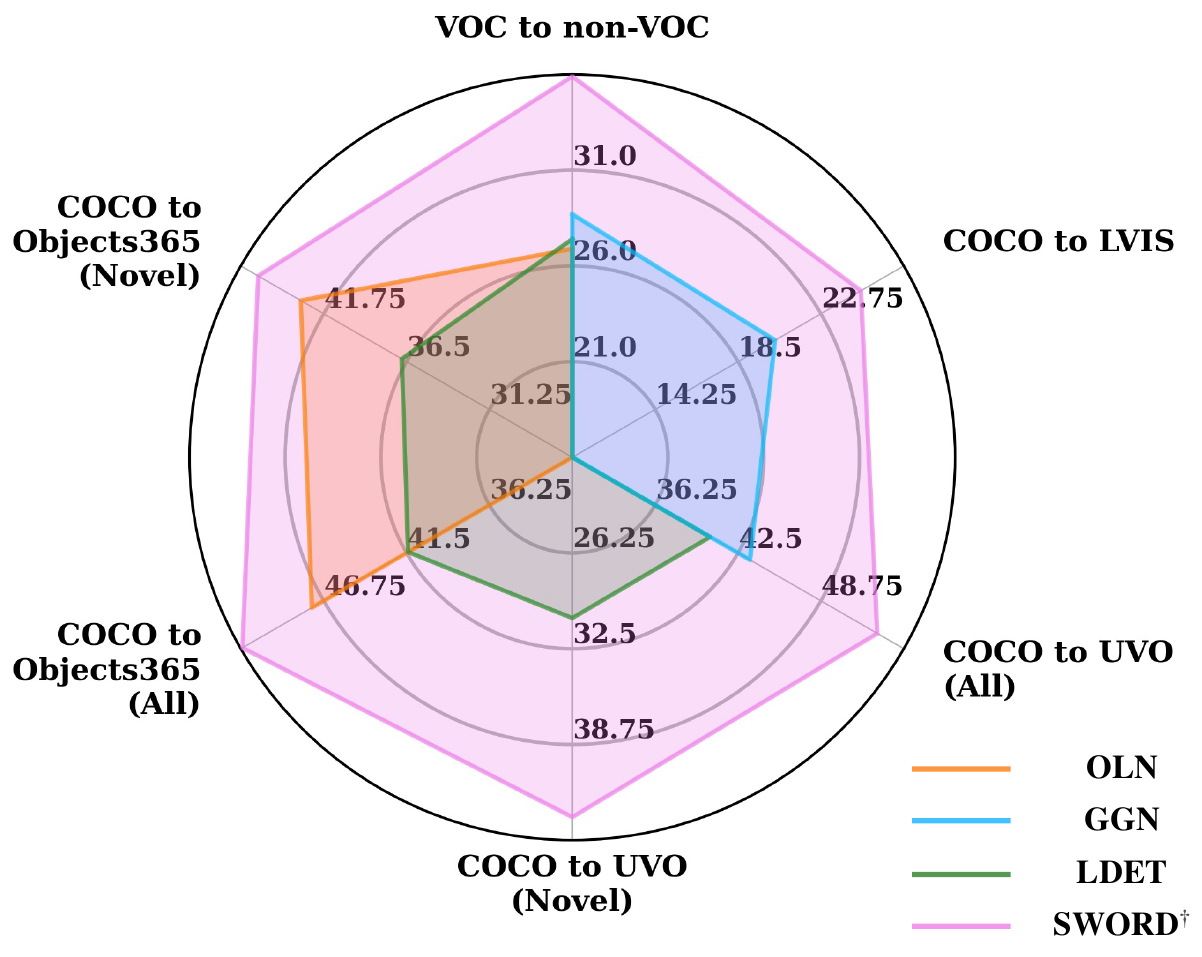}
\vspace{-6mm}
\caption{\modelname{}$^{\dag}$ achieves the state-of-the-art performance on various settings compared with other open-world methods. The results are reported based on AR$^{\rm m}_{100}$ by default. The metric of Objects365~\cite{shao2019objects365} is AR$^{\rm b}_{100}$ since it does not provide mask annotations.}
\label{fig:radar}
\vspace{-5mm}
\end{figure}

In this work, we cut off the above obstacles and propose \modelname{}, unsealing the \underline{s}ecrets of Transformer-based models for open-\underline{wor}l\underline{d} instance segmentation. We first introduce to attach a \texttt{stop-grad} operation before the classification head and further add the IoU heads for predicting object scores. This not only prevents the \textit{novel} object from being suppressed as background so as to improve the recall ability of network, but also allows the DETR-like models to preserve the classification head for heuristic label assignment. Using \texttt{stop-grad} alone, however, would inevitably reduce the network's discrimination. Therefore, we then design a novel contrastive learning framework for learning the discriminative representations between objects and background. The core idea is to ensure similar representations among objects while enlarging the distinction between the objects and background in the feature space. Specifically, we maintain a universal object queue to store the annotated object embeddings during training. The averaged feature of the queue, \textit{i.e.}, object center, captures the common characteristics of objects and plays as the role of \textit{query} in contrastive learning. The positive and hard negative samples are dynamically selected from query embeddings according to the matching cost~\cite{villani2009optimal, peyre2019computational} with ground-truth. 
Contrastive learning is the key to reducing false positives for network and greatly improves average precision.

To this end, \modelname{} not only reveals the strong ability in recalling \textit{novel} objects, but also achieves high average precision. We further develop a model \modelname{}$^{\dag}$, which has exactly the same architecture of Deformable-DETR~\cite{zhu2020deformable-detr} but is trained on the combination of annotations and pseudo ground-truths generated by \modelname{}. \modelname{}$^{\dag}$ shows clear performance gains compared with its counterpart under the open-world setups. To summarize, the contributions of this work are:

\vspace{-2mm}
\begin{itemize}[leftmargin=*]
\item A simple yet effective framework \modelname{} is presented, which is the first study of Transformer-based model for open-world instance segmentation.
\vspace{-2mm}
\item We introduce the \texttt{stop-grad} operation to kill two birds with one stone: preventing the side-effect of classification learning to discover \textit{novel} objects in open-world setups, and enabling the heuristic label assignment for DETR-like models. 
\vspace{-2mm}
\item We design a novel contrastive learning method to learn the discriminative representations between objects and background, which is essential for achieving high average precision for the network.
\vspace{-2mm}
\item Extensive experiments demonstrate that our models achieve state-of-the-art performance on several benchmarks including COCO~\cite{lin2014coco}, LVIS~\cite{gupta2019lvis}, UVO~\cite{wang2021uvo} and Objects365~\cite{shao2019objects365}, as shown in Figure~\ref{fig:radar}.
\end{itemize} 







\vspace{-1mm}
\section{Related Work}\label{sec:related_work}
\vspace{-1mm}

\myparagraph{Open-world Instance Segmentation.} Towards building more practical applications in the real world, the open-world-related problems~\cite{bendale2015towards, cen2021deep, han2019learning, cen2021deep, qi2021open, kwon2020backpropagated, sun2022gradient, yang2021objects, lin2022vldet, yan2023uninext} have raised great attention recently. Kim et al.~\cite{kim2022oln} firstly establish the protocol of open-world instance segmentation. Literally, open-world models should not only segment all the previously seen objects, but also localize the unseen objects during inference.
There are several works~\cite{kim2022oln, saito2021ldet, wang2022ggn, xue2022single, huang2022good, wang2023openinst, kalluri2023udos} attempting to solve the problem from various aspects. We refer the readers to Appendix \textcolor{red}{A} for a comprehensive review of these works.

\begin{figure}[t]
\begin{center}
\includegraphics[width=1.0\linewidth]{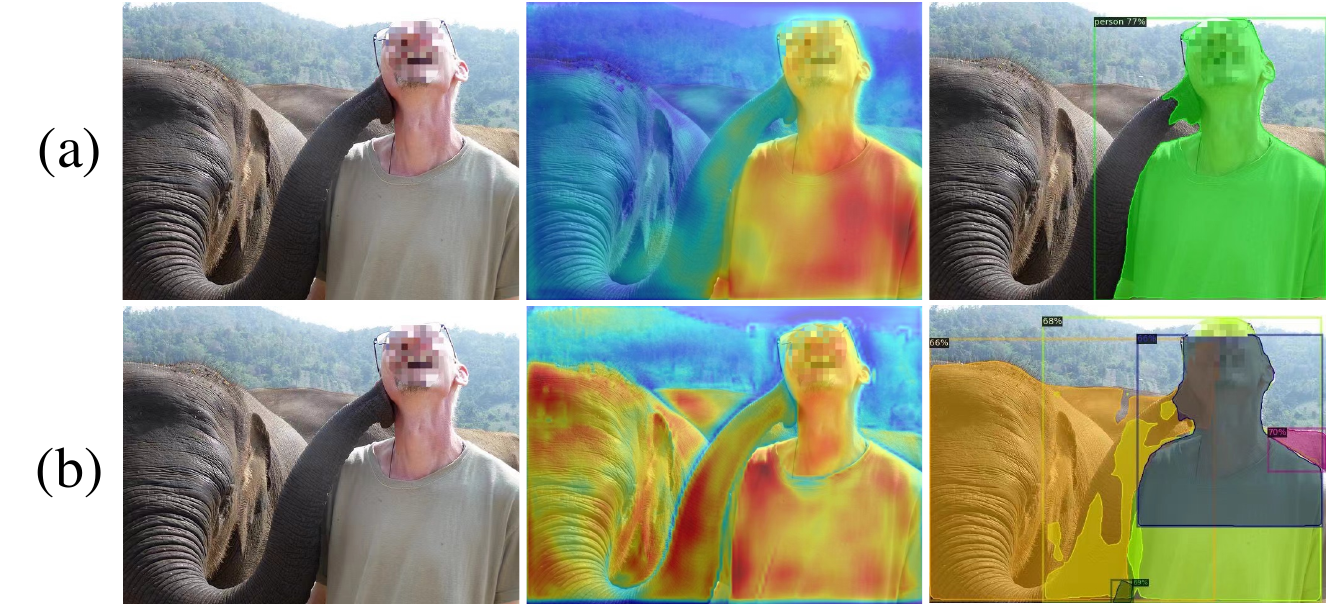}
\end{center}
\vspace{-4mm}
\caption{\textbf{Visualization results of (a) closed-world model Deformable-DETR and (b) open-world model OLN}. For each example, we show the input images, feature maps and predicted results from left to right. Note that \textit{elephant} is the unseen category in the training set.}
\label{fig:demo}
\vspace{-5mm}
\end{figure}

\myparagraph{\emph{Discussion.}} To dig out the problems of current models in open-world scenario, we visualize the predicted results of a closed-world model Deformable-DETR~\cite{zhu2020deformable-detr} and an open-world model OLN~\cite{kim2022oln} in Figure~\ref{fig:demo}. (i) For Deforamble-DETR, we notice that the \textit{novel} objects have low activations in the feature map, which we term as \textit{feature degradation}. This is because \textit{novel} objects are treated as background during training. As shown in the middle picture of Figure~\ref{fig:demo}\textcolor{red}{a}, the elephant is the unseen category and its feature can hardly be distinguished from its surroundings. (ii) OLN~\cite{kim2022oln} has the generalized ability to localize \textit{novel} objects. Nevertheless, it suffers from the problem of producing many false positives, \textit{e.g.}, parts of the man's body in Figure~\ref{fig:demo}\textcolor{red}{b}. OLN~\cite{kim2022oln} is trained with positive samples and can not perceive backdrop. It would assign high scores for all the object proposals. To conclude, there are two critical issues for the open-world instance segmentation: \emph{preventing the \textit{feature degradation} of \textit{novel} objects} and \emph{learning the discrimination between objects and background}.

\myparagraph{Contrastive Learning.} Self-supervised learning could be divided into three groups: contrastive learning, self-distillation~\cite{caron2021dino, zhou2021ibot} and masked image modeling~\cite{he2022mae, chen2022cae, tian2023spark}. Among which, contrastive learning~\cite{he2020moco, chen2020simclr, grill2020byol, xie2021detco, wang2021densecl, khosla2020supcontrast, bai2022region, khosla2020supcontrast, han2020coclr} has been dominant for a long time. The core idea of it lies in that the positive samples are attracted while the negative samples are pulled away in the feature space to learn the discriminative representations. MoCo~\cite{he2020moco} maintains a memory queue to store a large number of negative pairs and enables the momentum update of the memory encoder to guarantee the queue feature consistency. MoCo v3~\cite{chen2021mocov3} discards the memory queue and conducts self-supervised training on visual Transformers. SimSiam~\cite{chen2021simsiam} develops the extremely simple siamese network without any negative sample, and points out a stop-gradient operation plays an essential role in preventing mode collapse. In this work, we absorb the ideas from contrastive learning to learn the distinct representations of objects and background.

\vspace{-1mm}
\section{Method}\label{sec:method}
\vspace{-1mm}

\begin{figure*}[t]
\begin{center}
\includegraphics[width=0.98\textwidth]{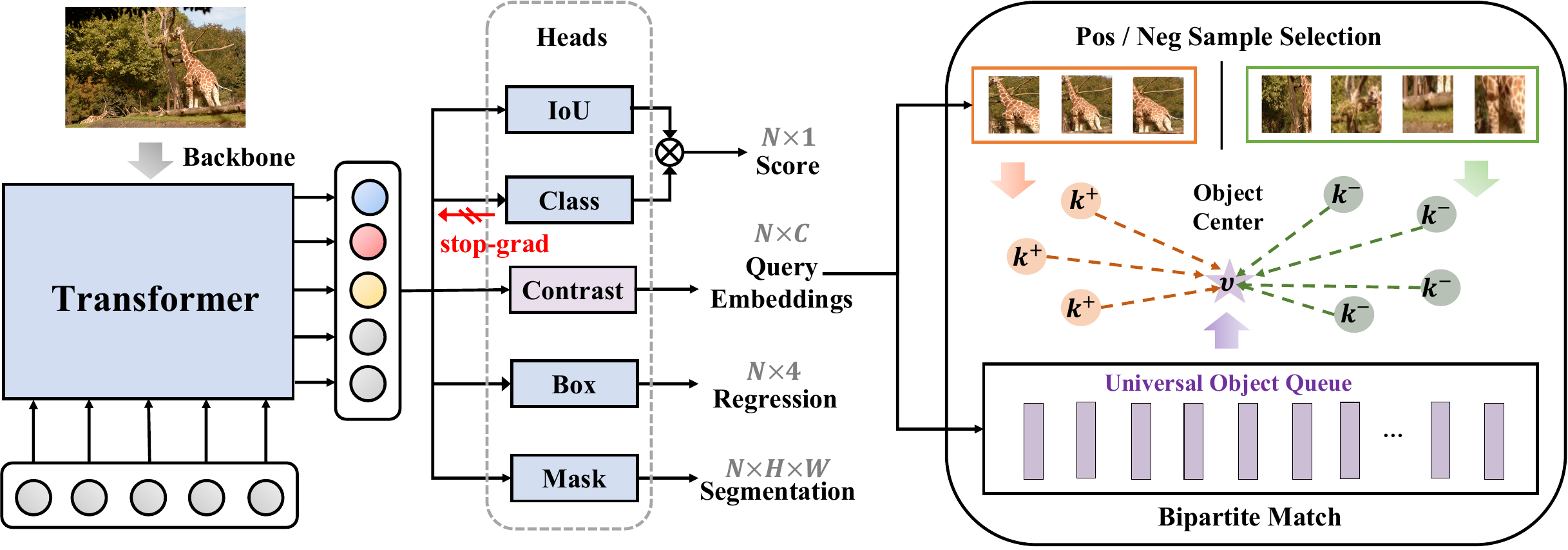}
\end{center}
\vspace{-2mm}
\caption{\textbf{The overall framework of SWORD}. We first attach a \texttt{stop-grad} operation before the classification head and  add the IoU head to help discover \textit{novel} objects. Further, a contrastive head is added on top of the Transformer decoder to predict the query embeddings for contrastive learning. It is essential for the network to learn distinct representations between objects and background. During inference, the classification scores and IoU scores are fused to produce the final scores.}
\label{fig:main}
\vspace{-3mm}
\end{figure*}

\subsection{Overview} \label{subsec:definition}

Open-world instance segmentation problem introduced by recent advanced works~\cite{kim2022oln, saito2021ldet, wang2022ggn} can be formulated as follows. Given an instance segmentation dataset (\textit{e.g.}, COCO with 80 classes), we have the object annotations on the \textit{base} category set $\mathcal{C}_{\text{base}}$ (\textit{e.g.}, $\mathcal{C}_{\text{base}}$ contains 20 PASCAL-VOC classes). Notably, there are also a large number of \textit{novel} objects co-appearing in the images while remaining un-annotated. The models are trained with the base-category annotations to provide a set of class-agnostic proposals $\mathcal{P} = \left \{ s_{i}, b_{i}, m_{i} \right \}_{i=1}^{p}$ to localize \textit{all} objects in the image, where $s_{i} \in \mathbb{R}$ indicates the proposal score, $b_{i} \in \mathbb{R}^{4}$ denotes the bounding box coordinates and $m_{i} \in \mathbb{R}^{H \times W}$ is the segmentation mask for the $i$-th prediction. The generalization of models is evaluated by segmenting the \textit{novel} objects from the unseen category set $\mathcal{C}_{\text{novel}}$ (\textit{e.g.}, 60 non-PASCAL-VOC classes) in a class-agnostic fashion.

The overall framework of our proposed \modelname{} is illustrated in Figure~\ref{fig:main}. Our network is based on the closed-world model Deformable-DETR~\cite{zhu2020deformable-detr}, and we explain how to convert it into an open-world instance segmentation model in Section~\ref{subsec:close2open}. We then propose to utilize contrastive learning to generate distinct representations for objects and background in Section~\ref{subsec:sword}. Additionally, we introduce an extension of our model, denoted as \modelname{}$^{\dag}$, through pseudo labeling based self-training in Section~\ref{subsec:sword+}.

\subsection{Open-world Transformer} \label{subsec:close2open}

Our network is built upon the Deformable-DETR~\cite{zhu2020deformable-detr} due to its simple architecture. First, we add the mask head on top of the Transformer to generate the instance masks by performing dynamic convolution~\cite{tian2020condinst, cheng2022mask2former, wu2022idol}. Then, the model is transformed into an open-world model for learning class-agnostic mask proposals with following designs.

\begin{table}[t]
    \centering
    \renewcommand\arraystretch{1.05} 
    \setlength{\tabcolsep}{5.0pt}    
    \small
    \captionsetup{width=0.98\linewidth}
    \caption{The illustration of the key role of \texttt{stop-grad} operation. `class' and `IoU' in the first column mean classification head and IoU head, respectively.}
    \vspace{-2mm}



 






\begin{tabular}{l|cc}
\Xhline{1.0pt}

\multicolumn{1}{l|}{Variants} & 
\multicolumn{1}{c}{\begin{tabular}[c]{@{}c@{}}Network\\ Generalization\end{tabular}} &
\multicolumn{1}{c}{\begin{tabular}[c]{@{}c@{}}Heuristic \\ Label Assignment\end{tabular}}  \\

\hline

class only & \xmark & \cmark  \\
IoU only & \cmark & \xmark  \\
class + IoU & \xmark & \cmark  \\
class + IoU + stop-grad & \cmark & \cmark \\

\Xhline{1.0pt}
\end{tabular}

    \label{tab:stop-grad}
    \vspace{-5mm}
\end{table}

\myparagraph{IoU Heads.} Inspired by the philosophy of OLN~\cite{kim2022oln} that localization quality is a better objectness cue than the classification score in the open-world setting, we add the extra two IoU heads on top of the Transformer decoder to predict the box IoU score $c_{b}$ and mask IoU score $c_{m}$, respectively. The IoU scores are helpful to discover the \textit{novel} objects.

\myparagraph{Stop-gradient Operation.} The object-or-not learning of classification can hurt the generalization of network, while DETR-like models need to preserve the classification head for heuristic label assignment (\textit{e.g.}, Hungarian matching~\cite{carion2020detr}). \emph{This leads to a conflict situation}. To address this issue, we propose a simple yet effective solution by introducing a \texttt{stop-grad} operation~\cite{qiao2021defrcn, chen2021simsiam} before the classification head. This operation prevents the gradient passing from the classification head to the network, which avoids suppressing all un-annotated regions as background. On the other hand, it can be seamlessly applied on the advanced DETR-like detectors~\cite{carion2020detr, liu2022dab-detr, li2022dn-detr} to facilitate the heuristic label assignment. As illustrated in Table~\ref{tab:stop-grad}, the \texttt{stop-grad} operation can confer all the desired properties upon the DETR-like models, thus enabling the networks to possess the open-world capacity in a manner akin to the OLN paradigm.

\subsection{Contrastive Learning} \label{subsec:sword}

The use of \texttt{stop-grad} operation would also disable the discriminative ability of network. To address the issue, in this subsection, we propose a contrastive learning~\cite{he2020moco, grill2020byol, chen2021simsiam} framework to learn the distinct representations between objects and background.

\myparagraph{Universal Object Queue.} As shown in Figure~\ref{fig:main}, we further add the contrastive head on top of the Transformer decoder to learn the query embeddings. The parameters of the constrastive head are also copied to a momentum branch using the exponential moving average (EMA) method:

\vspace{-2mm}
\begin{equation}
    \theta_{c}^{'} \leftarrow \alpha\theta_{c}^{'} + (1 - \alpha) \theta_{c}
    \label{eq:ema}
\end{equation}

\noindent
where $\theta_{c}$ and $\theta_{c}^{'}$ denote the parameters of the regularly and EMA updated contrastive head, respectively. $\alpha$ is the momentum rate. We use a universal object queue $\mathcal{Q}$ to store the object embeddings, where each element is the projected feature from the momentum contrastive head. Given an image, we select those queries best matched to the ground-truth through bipartite matching and store their projected embeddings into $\mathcal{Q}$ using the first-in-first-out strategy. 
We then average all the features in the universal object queue to get the object center $v$. Intuitively, the object center captures the common object characteristics and stays stable in the feature space.

\myparagraph{Positive and Negative Samples.} It is well known that the positive and negative samples play the essential role in contrastive learning. Here, we dynamically select the positive and negative pairs according to a optimal transport assignment method~\cite{ge2021ota, villani2009optimal, peyre2019computational, ge2021yolox}. Specifically, given an image, we take the classification results into consideration and compute the costs between predictions and ground-truths:

\vspace{-2mm}
\begin{equation}
    \mathcal{C} = \lambda_{cls} \cdot \mathcal{C}_{cls} + \lambda_{L1} \cdot \mathcal{C}_{L1} + \lambda_{giou} \cdot \mathcal{C}_{giou}
    \label{eq:matching_cost}
\end{equation}

\noindent
where, $\mathcal{C}_{cls}$ is focal loss~\cite{lin2017focal}. The box-related losses include the $\mathcal{L}_{1}$ loss and generalized IoU loss~\cite{rezatofighi2019giou}. 
Ideally, the predictions with the least cost are those objects close to the ground-truths. To improve the quality of learned embeddings, for each ground-truth object, we first dynamically choose $k_{1}$ and $k_{2}$ predictions with the least cost, where $k_{2}>k_{1}$. Then the $k_{1}$ predictions are positive samples, and the left $k_{2}-k_{1}$ predictions are considered as hard negatives.

\myparagraph{Contrastive Loss.} In the contrastive learning, we expect the positive samples $\mathcal{K}^{+}$ should be close to the object center $v$ while the negative ones $\mathcal{K}^{-}$ should pulled away. The contrastive loss~\cite{he2020moco} is formulated as:

\begin{equation} 
    \mathcal{L}_{con} = -\text{log} \frac{\sum_{k^{+} \in \mathcal{K}^{+}} \text{exp}(v \cdot k^{+}) }{\sum_{k^{+} \in \mathcal{K}^{+}} \text{exp}(v \cdot k^{+}) + \sum_{k^{-} \in \mathcal{K}^{-}} \text{exp}(v \cdot k^{-})}
    \label{eq:loss_contrast}
\end{equation}

\subsection{Training and Inference}

\myparagraph{Training.} The label assignment for the network optimization also relies on the matching cost as Eq. (\ref{eq:matching_cost}). The predictions with the least costs are assigned to ground-truths as positive samples and others as negatives~\cite{ge2021yolox}. The overall loss function for training is:

\vspace{-4mm}
\begin{equation}
\begin{aligned}
    \mathcal{L} &= \lambda_{cls} \cdot \mathcal{L}_{cls} + \lambda_{L1} \cdot \mathcal{L}_{L1} + \lambda_{giou} \cdot \mathcal{L}_{giou} \\
    &+ \lambda_{mask} \cdot \mathcal{L}_{mask} + \lambda_{dice} \cdot \mathcal{L}_{dice} \\
    &+ \lambda_{iou} \cdot \mathcal{L}_{iou} + \lambda_{con} \cdot \mathcal{L}_{con} 
\end{aligned}
\label{eq:loss}
\end{equation}
\vspace{-2mm}

\noindent
where, $\mathcal{L}_{cls}$, $\mathcal{L}_{L1}$ and $\mathcal{L}_{giou}$ are the same as the components in Eq.~\ref{eq:matching_cost}. The mask-related loss is a combination of the mask binary loss and DICE loss ~\cite{milletari2016vnet}. The IoU scores are supervised by the binary cross entropy loss $\mathcal{L}_{iou}$.

\myparagraph{Inference.} We use the geometric mean of classification scores and IoU scores as final scores, \textit{i.e.}, $s = \sqrt[3]{c_{c} \cdot c_{b} \cdot c_{m}}$. And the top-100 predictions are left for evaluation.

\subsection{Extension: Pseudo Ground-truth Training} \label{subsec:sword+}

The advanced approach GGN~\cite{wang2022ggn} proves that using the pseudo ground-truth for training can significantly boost the performance of Mask-RCNN~\cite{he2017maskrcnn} in the open-world setup. Therefore, we also adopt the pseudo labeling method and develop an extension model, \modelname{}$^{\dag}$. Following the existing practice of GGN ~\cite{wang2022ggn}, we use \modelname{} as teacher model to generate the pseudo boxes/masks. After filtering out those predictions having high box overlap with ground-truth, we add the remaining top-$k$ predictions to the ground-truth annotations. Finally, the standard Deformable-DETR is trained under the supervision of augmented annotations. 
Please see more details in Appendix \textcolor{red}{B}.

\vspace{-1mm}
\section{Experiments}\label{sec:experiments}
\vspace{-1mm}

In this section, we first thoroughly evaluate the performance of proposed models in two challenging open-world settings: cross-category and cross-dataset generalizations. Then we conduct extensive ablation studies to discuss the key designs and analyze the crucial issues in Sec.~\ref{subsec:ablation}.

\vspace{-1mm}
\subsection{Experiment Settings}
\vspace{-2mm}

\myparagraph{Datasets.} Our experiments are conducted on COCO~\cite{lin2014coco}, LVIS~\cite{gupta2019lvis}, UVO~\cite{wang2021uvo} and Objects365~\cite{shao2019objects365} datasets. \textbf{COCO} is the widely used instance segmentation benchmark with 80 categories. \textbf{LVIS} shares the same images with COCO while having a more complete label system. It has a large taxonomy of 1203 categories in a long-tailed distribution. \textbf{UVO} originates from the Kinetics400~\cite{carreira2017k400} dataset and all the instance masks are exhaustively annotated. \textbf{Objects365} is a large-scale object detection dataset with 365 categories where all the COCO 80 categories are included.

We target at two challenging open-world generalization setups~\cite{kim2022oln, saito2021ldet, wang2022ggn}: \textbf{(1) Cross-category generalization}. 
On COCO benchmark, we follow the common practice~\cite{kim2022oln, saito2021ldet} to split the annotations into two non-overlapping class sets, where the PASCAL-VOC~\cite{everingham2010pascal} 20 classes are adopted as the \textit{base} set and the rest of 60 non-VOC classes are \textit{novel} set. The second benchmark, \textit{i.e.}, LVIS, splits the 1203 classes into 80 COCO classes for training and the remaining 1123 non-COCO classes for evaluation. For cross-category generalization, the results are reported on the unseen categories. \textbf{(2) Cross-dataset generalization}. This is to evaluate the model's open-world generalization ability when used in the wild. COCO is used as the training source and the models are tested on new datasets, \textit{i.e.}, UVO \footnote{The downsampled dense split of v1.0 contains two classes: ``objects" for COCO categories and ``other" for non-COCO categories. The NOVEL metrics are measured on the ``other'' categories. The previously released v0.5 does not distinguish the object categories and all the objects are annotated as ``objects''. We report the results of ALL metrics based on this version following the previous works~\cite{saito2021ldet, wang2022ggn}} and Objects365\footnote{Objects365 only has box annotations, so we report the results regarding the box metrics.}. In this setting, we show the results on both \textit{novel} categories and all categories (including base and novel ones).

\myparagraph{Evaluation Metrics.} Following previous works~\cite{kim2022oln, saito2021ldet, wang2022ggn}, we use average recall (AR@k) and average precision (AP) over multiple IoU thresholds $\left [ 0.5:0.95 \right ]$ to measure the performance. The proposal number k is set as 100 by default. The superscripts `$\rm b$' and `$\rm m$' denote the boxes and masks, respectively. And AR$_{s/m/l}$ represent AR@100 for small, medium and large size of objects. Notably, \textbf{the most concerned metric} in open-world scenario~\cite{kim2022oln} is \textbf{AR@100}.

\myparagraph{Implementation Details.} 
In all setups, models are trained and evaluated in a class-agnostic way. We use ResNet-50~\cite{he2016resnet} as backbone by default and the Transformer network has 6 encoders and 6 decoders with the hidden dimension of 256. We use 2000 object queries when the training source is VOC(COCO), otherwise the query number is set as 1000. The size of the universal object queue is set as 4096 and the EMA rate $\alpha$ is 0.999. The values of $k_{1}$ and $k_{2}$ for contrastive learning are set as 10 and 100, respectively. In all our experiments, we also train the Deformable-DETR for comparison using the same setting as ours (\emph{i.e.}, object query number, training epochs) for fair comparisons. Please see Appendix \textcolor{red}{C} for more implementation details.

\vspace{-1mm}
\subsection{Cross-category Generalization} \label{subsec:cross-category}
\vspace{-2mm}

\myparagraph{VOC to non-VOC.} 
In Table~\ref{tab:voc}, we compare our methods with other state-of-the-art methods in VOC to non-VOC setup. It shows that \modelname{} yields the significant 7.9\% gain on AR$_{100}^{\rm b}$ and 7.5\% on AR$_{100}^{\rm m}$ compared with the Deformable-DETR baseline. And our model outperforms all the previous single model. 
The performance could be further boosted by exploiting the pseudo ground-truth of \modelname{}. \modelname{}$^{\dag}$ achieves state-of-the-art performance in all metrics, \textit{e.g.}, 40.0\% on AR$_{100}^{\rm b}$ and 34.9\% on AR$_{100}^{\rm m}$.

\myparagraph{COCO to LVIS.} Table~\ref{tab:lvis} summarizes the performance of different methods in the COCO to LVIS setup. Compared to the Deformable-DETR baseline, \modelname{} shows an obvious performance gain, with +4.1\% on AR${100}^{\rm b}$ and +4.0\% on AR${100}^{\rm m}$. Additionally, \modelname{}$^{\dag}$ outperforms the previous best method GGN~\cite{wang2022ggn} by 5.6\% AR$_{100}^{\rm b}$, which is a relative improvement of 25\%.

\begin{table}[t]
\centering
\renewcommand\arraystretch{1.05} 
\setlength{\tabcolsep}{1.0pt}    
\small
\caption{\small{State-of-the-art performance in VOC to non-VOC setup.}}
\vspace{-2mm}
\begin{tabular}{l|ccc|ccc}
\Xhline{1.0pt}


Method & AP$^{\rm b}$ & AR$_{10}^{\rm b}$ & AR$_{100}^{\rm b}$ &
         AP$^{\rm m}$ & AR$_{10}^{\rm m}$ & AR$_{100}^{\rm m}$ \\
\Xhline{1.0pt}

Mask-RCNN~\cite{he2017maskrcnn} & 1.6 & 10.2 & 23.5 & 0.9 & 7.9 & 17.7 \\
OLN~\cite{kim2022oln}       & 3.7 & 18.0 & 33.5 & -   & 16.9   & -    \\
LDET~\cite{saito2021ldet}      & 5.0 & 18.2 & 30.8 & 4.3 & 16.3 & 27.4 \\
GGN~\cite{wang2022ggn}       & 5.8 & 17.3 & 31.6 & 4.9 & 16.1 & 28.7 \\
GGN + OLN~\cite{wang2022ggn} & 3.4 & 17.1 & 37.2 & 3.2 & 16.4 & 33.7 \\
Deformable-DETR~\cite{zhu2020deformable-detr} & 2.5 & 12.2 & 27.4 & 2.2 & 10.2 & 22.7 \\

\hline

\rowcolor{lightgray!28}\modelname{} (Ours) & 5.8 & 17.8 & 35.3 & 4.8 & 15.7 & 30.2 \\
\rowcolor{lightgray!28}\modelname{}$^{\dag}$ (Ours) & \textbf{6.2} & \textbf{22.0} & \textbf{40.0} & \textbf{5.8} & \textbf{20.2} & \textbf{34.9} \\

\Xhline{1.0pt}
\end{tabular}
\label{tab:voc} 
\vspace{-2mm}
\end{table}

\begin{table}[t]
\centering
\renewcommand\arraystretch{1.05} 
\setlength{\tabcolsep}{4.5pt}    
\small
\caption{\small{State-of-the-art performance in COCO to LVIS setup.}}
\vspace{-2mm}
\begin{tabular}{l|cc|cc}
\Xhline{1.0pt}


Method & AR$_{10}^{\rm b}$ & AR$_{100}^{\rm b}$ &
         AR$_{10}^{\rm m}$ & AR$_{100}^{\rm m}$ \\
\Xhline{1.0pt}

Mask-RCNN~\cite{he2017maskrcnn} & 6.1 & 19.4 & 5.6 & 17.2 \\
GGN~\cite{wang2022ggn}       & 7.6 & 22.4 & 7.2   & 20.4 \\
Deformable-DETR~\cite{zhu2020deformable-detr} & 6.3 & 19.4 & 5.5 & 16.4 \\

\hline

\rowcolor{lightgray!28}\modelname{} (Ours) & 8.8 & 23.5 & 8.0 & 20.4 \\
\rowcolor{lightgray!28}\modelname{}$^{\dag}$ (Ours) & \textbf{9.8} & \textbf{28.0} & \textbf{9.0} & \textbf{23.8} \\

\Xhline{1.0pt}
\end{tabular}




\label{tab:lvis} 
\vspace{-4mm}
\end{table}

\begin{table*}[t]
\centering
\renewcommand\arraystretch{1.05} 
\setlength{\tabcolsep}{3.5pt}    
\small
\caption{Comparison of state-of-the-art performance in COCO to UVO setup. Top rows: Models are trained with 20 PASCAL-VOC classes on COCO dataset. Bottom rows: Models are trained with all 80 COCO classes on COCO dataset.}
\vspace{-2mm}





\begin{tabular}{l|c|ccc|ccc| ccc|ccc}
\Xhline{1.0pt}

\multirow{2}{*}{Method} & \multirow{2}{*}{Train} & \multicolumn{6}{c|}{Novel}  & \multicolumn{6}{c}{All} \\
\cline{3-14} 
& & AP$^{\rm b}$ & AR$_{10}^{\rm b}$ & AR$_{100}^{\rm b}$
 & AP$^{\rm m}$ & AR$_{10}^{\rm m}$ & AR$_{100}^{\rm m}$ 
 & AP$^{\rm b}$ & AR$_{10}^{\rm b}$ & AR$_{100}^{\rm b}$ 
 & AP$^{\rm m}$ & AR$_{10}^{\rm m}$ & AR$_{100}^{\rm m}$ \\
\Xhline{1.0pt}

Mask-RCNN~\cite{he2017maskrcnn} & \multirow{6}{*}{\begin{tabular}[c]{@{}c@{}}VOC \\ (COCO)\end{tabular}} & 5.9 & 11.4 & 16.2 & 2.3 & 7.6 & 11.4 & 20.2 & 25.3 & 30.8 & 15.7 & 20.1 & 24.3 \\
LDET~\cite{saito2021ldet}  &  & 9.3 & 16.0 & 31.9 & 4.9 & 12.3 & 25.2 & 22.7 & 28.1 & 43.3 & 18.7 & 23.9 & 36.0 \\
Deformable-DETR~\cite{zhu2020deformable-detr} &
          & 7.2 & 13.5 & 33.5 & 3.4 & 9.5 & 25.3 & 23.4 & 29.4 & 49.8 & 19.1 & 24.0 & 39.4 \\
\rowcolor{lightgray!28}\modelname{} (Ours) &
          & 11.2 & 16.8 & 43.1 & 6.1 & 13.3 & 34.9 & \textbf{24.9} & 30.6 & 55.3 & 19.6 & 25.3 & 45.2 \\
\rowcolor{lightgray!28}\modelname{}$^{\dag}$ (Ours) &
          & \textbf{11.8} & \textbf{18.4} & \textbf{45.6} & \textbf{8.4} & \textbf{16.8} & \textbf{38.1} & 23.4 & \textbf{31.1} & \textbf{59.2} & \textbf{21.0} & \textbf{28.4} & \textbf{49.5} \\
          
\hline

Mask-RCNN~\cite{he2017maskrcnn} & \multirow{6}{*}{COCO} & 11.8 & 16.4 & 30.4 & 7.0 & 13.8 & 25.5 & 25.7 & 30.2 & 43.8 & 20.7 & 25.7 & 36.7 \\
LDET~\cite{saito2021ldet}   &  & 12.9 & 19.0 & 35.9 & 8.2 & 15.9 & 30.5 & 26.0 & 30.9 & 47.0 & 22.1 & 27.3 & 40.7 \\
GGN~\cite{wang2022ggn}    &  & -    & -    & -    & -   & -    & -    & 24.0 & 29.8 & 52.2 & 20.3 & -    & 43.4 \\
Deformable-DETR~\cite{zhu2020deformable-detr} &
          & 14.2 & 20.0 & 45.8 & 9.0 & 16.7 & 37.9 & 29.1 & 35.0 & 60.7 & 24.7 & 30.1 & 50.3 \\

\rowcolor{lightgray!28}\modelname{} (Ours) &
          & \textbf{17.5} & 22.2 & 48.1 & \textbf{12.8} & 19.4 & 40.6 & \textbf{32.0} & \textbf{36.5} & 61.2 & \textbf{28.0} & 32.4 & 51.5\\
\rowcolor{lightgray!28}\modelname{}$^{\dag}$ (Ours) &
          & 16.6 & \textbf{22.7} & \textbf{50.0} & 12.7 & \textbf{20.9} & \textbf{42.8} & 28.1 & 35.2 & \textbf{62.0} & 25.7 & \textbf{32.5} & \textbf{53.0} \\

\Xhline{1.0pt}
\end{tabular}
\label{tab:coco2uvo} 
\vspace{0mm}
\end{table*}

\begin{table*}[t]
\centering
\renewcommand\arraystretch{1.05} 
\setlength{\tabcolsep}{5.8pt}    
\small
\caption{Comparison of state-of-the-art performance in COCO to Objects365 setup.}
\vspace{-2mm}
\begin{tabular}{l|cccccc|cccccc}
\Xhline{1.0pt}

\multirow{2}{*}{Method} & \multicolumn{6}{c|}{Novel}  & \multicolumn{6}{c}{All} \\
\cline{2-13} & 
  AP$^{\rm b}$ & AR$_{10}^{\rm b}$ & AR$_{100}^{\rm b}$ & AR$_{s}^{\rm b}$ & AR$_{m}^{\rm b}$ & AR$_{l}^{\rm b}$ &
  AP$^{\rm b}$ & AR$_{10}^{\rm b}$ & AR$_{100}^{\rm b}$ & AR$_{s}^{\rm b}$ & AR$_{m}^{\rm b}$ & AR$_{l}^{\rm b}$ \\

\Xhline{1.0pt}

Mask-RCNN~\cite{he2017maskrcnn} & 13.0 & 19.3 & 32.8 & 18.2 & 36.4 & 43.5 & 25.1 & 23.9 & 40.3 & 22.7 & 42.8 & 53.4\\
LDET~\cite{saito2021ldet}      & 12.8 & 20.0 & 36.8 & 20.7 & 40.5 & 48.9 & 22.5 & 22.7 & 41.4 & 22.9 & 44.3 & 54.9\\
Deformable-DETR~\cite{zhu2020deformable-detr}
          & 12.9 & 19.0 & 40.1 & 22.8 & 43.4 & 54.1 & 27.3 & 25.3 & 48.7 & 27.5 & 50.9 & 65.6\\
\hline

\rowcolor{lightgray!28}\modelname{} (Ours) 
          & \textbf{16.6} & 22.8 & 43.9 & 25.0 & 48.6 & 57.6 & \textbf{29.7} & \textbf{27.3} & 50.8 & 28.6 & 54.0 & 67.2 \\
\rowcolor{lightgray!28}\modelname{}$^{\dag}$ (Ours)
          & 16.3 & \textbf{23.5} & \textbf{45.9} & \textbf{25.9} & \textbf{50.5} & \textbf{60.7} & 28.7 & 27.2 & \textbf{51.9} & \textbf{29.4} & \textbf{55.4} & \textbf{68.4} \\

\Xhline{1.0pt}
\end{tabular}

\label{tab:coco2obj365} 
\vspace{-2mm}
\end{table*}

\vspace{-1mm}
\subsection{Cross-dataset Generalization} \label{subsec:cross-dataset}
\vspace{-2mm}

\myparagraph{COCO to UVO.} 
For COCO to UVO generalization, we evaluate models trained with 20 PASCAL-VOC classes and all 80 COCO classes. Table~\ref{tab:coco2uvo} presents a thorough comparison of results for both \textit{novel} and \textit{all} objects. Deformable-DETR performs considerable well in this setting, outperforming previous methods by a large margin. Our proposed model, \modelname{}, further improved the performance over the strong baseline for all metrics. Notably, in the partially annotated VOC(COCO) to UVO scenario, \modelname{} demonstrated a significant 9.6\% AR$^{\rm m}_{100}$ gain for \textit{novel} objects. This highlights the effectiveness of our model in discovering \textit{novel} objects.

Another observation is that pseudo ground-truth training is effective in improving the average recall (AR) of models but may decrease the average precision (AP). By comparing Deformable-DETR and \modelname{}$^{\dag}$ (both have the same architecture), we noticed a consistent gain in AR. However, AP$^{\rm b}$ for all objects dropped from 29.1\% to 28.1\% in the last rwo. The reason may attribute to that false positives in the pseudo ground-truth labels can misguide the model training.

\myparagraph{COCO to Objects365.} The results of COCO to Objects365 generalization are listed in Table~\ref{tab:coco2obj365}. Mask-RCNN-based method LDET\footnote{We report the results of LDET~\cite{saito2021ldet} using the same class-agnostic evaluation for fair comparison, whereas results in the original paper are based on the class-wise evaluation.} improves Mask-RCNN in terms of ARs but decrease APs. 
\modelname{} significantly outperforms Deformable-DETR baseline for all metrics and achieve state-of-the-art performance.

\vspace{-1mm}
\subsection{Ablation Study} \label{subsec:ablation}
\vspace{-1mm}

In this subsection, we conduct extensive ablation studies to analyze the crucial composing of our method. The experimental results are based on the COCO (80 classes) to UVO setting and the backbone is ResNet50 otherwise specified. We report the result in terms of mask metrics.

\myparagraph{Analysis of Key Designs.} Table~\ref{tab:ab_component} presents the ablation results to study the key designs of our method. In addition to the standard Deformable-DETR, we also build an OLN version Deformable-DETR by replacing the classification head with IoU heads to make a comparison. Since COCO is fully annotated and there are hardly any \textit{novel} objects, the co-existence of objects and background in un-annotated areas is not a significant issue. Therefore, we mainly focus on the discussion of \textbf{VOC(COCO) to UVO generalization}.

Starting from the closed-world Deformable-DETR, we first introduce the \texttt{stop-grad} operation to transform it into an open-world model. This simple yet effective operation prevents un-annotated regions from being suppressed as background. The obvious background-to-foreground transition for \textit{novel} objects leads to the simultaneous improvement of AP and AR for \textit{novel} objects.  
However, the use of \texttt{stop-grad} reduces the network's discrimination, resulting in many false positives predictions and a performance drop in AP for \textit{all} objects. 
Henceforth, the proposed contrastive learning framework is indispensible. This design significantly improves AP$^{\rm m}$ for \textit{all} objects by 6.5\%, and all ARs show steady performance improvement. The two key designs are also validated in the COCO to UVO setup, where \texttt{stop-grad} improves AR and contrastive learning greatly increases AP. These results demonstrate the effectiveness of our proposed approach. 

From another perspective, our proposed \modelname{} also demonstrates performance advantages over the OLN-version Deformable-DETR. \modelname{} not only reveals consistent performance advantages on ARs, but also shows  5.1  and 11.2 points gain on the AP$^{\rm m}$ for VOC to UVO and COCO to UVO setups, respectively. These results proves the superiority of our method.

\begin{table*}[t]
\centering
\renewcommand\arraystretch{1.10} 
\setlength{\tabcolsep}{10.0pt}    
\small
\captionsetup{width=0.97\linewidth}
\caption{Ablation on the key designs of our method. 'D-DETR' represents Deformable-DETR~\cite{zhu2020deformable-detr}. The results of an OLN version Deformable-DETR are also presented for comparison. 
We start from the Deformable-DETR baseline and gradually add the key components. The final model is the proposed \modelname{}.}
\vspace{-2mm}
\begin{tabular}{l|lll|lll}
\Xhline{1.0pt}

\multirow{2}{*}{Variants} & \multicolumn{3}{c|}{Novel}  & \multicolumn{3}{c}{All} \\
\cline{2-7} 
 & \multicolumn{1}{c}{AP$^{\rm m}$} & \multicolumn{1}{c}{AR$_{10}^{\rm m}$} & \multicolumn{1}{c|}{AR$_{100}^{\rm m}$} 
 & \multicolumn{1}{c}{AP$^{\rm m}$} & \multicolumn{1}{c}{AR$_{10}^{\rm m}$} & \multicolumn{1}{c}{AR$_{100}^{\rm m}$} \\

\tabbl
\multicolumn{7}{c}{\emph{VOC(COCO) to UVO}}  \\

\textcolor{gray}{D-DETR OLN-version} & \textcolor{gray}{5.8} & \textcolor{gray}{11.4} & \textcolor{gray}{31.0} & \textcolor{gray}{14.5} & \textcolor{gray}{23.4} & \textcolor{gray}{43.1} \\

D-DETR 
          & 3.4 & 9.5 & 25.3 & 19.1 & 24.0 & 39.4 \\
          

          
+ stop-grad 
          & 4.7\posval{1.3} & 11.6\posval{2.1} & 34.1\posval{8.8} &
          13.1\negval{6.0} & 22.1\negval{1.9} & 45.1\posval{5.7} \\

+ contrastive learning
          & 6.1\posval{1.4} & 13.3\posval{1.7} & 34.9\posval{0.8} &
          19.6\posval{6.5} & 25.3\posval{3.2} & 45.2\posval{0.1} \\

\tabbl
\multicolumn{7}{c}{\emph{COCO to UVO}}  \\

\textcolor{gray}{D-DETR OLN-verision} & \textcolor{gray}{10.0} & \textcolor{gray}{17.5} & \textcolor{gray}{40.9} & \textcolor{gray}{17.8} & \textcolor{gray}{28.0} & \textcolor{gray}{51.2} \\

D-DETR 
          & 9.0 & 16.7 & 37.9 & 24.7 & 30.1 & 50.3 \\

+ stop-grad
          & 10.3\posval{1.3} & 18.4\posval{1.7} & 41.6\posval{3.7} & 21.6\negval{3.1} & 30.5\posval{0.4} & 52.0\posval{1.7} \\

+ contrastive learning 
          & 12.8\posval{2.5} & 19.4\posval{1.0} & 40.6\negval{1.0} & 28.0\posval{6.4} & 32.4\posval{1.9} & 51.5\negval{0.5}\\

\Xhline{1.0pt}
\end{tabular}




          

          


\label{tab:ab_component} 
\vspace{-5mm}
\end{table*}

\begin{figure}[t]
\centering
\begin{minipage}[t]{0.48\textwidth}
    \vspace{4mm}
    \centering
    \captionsetup{width=0.95\linewidth}
    \captionof{table}{Ablation on the classification cost for sample selection in contrastive learning.}
    \vspace{-2mm}
    \setlength{\tabcolsep}{1.5mm}{
    \scalebox{1.0}{
    \small




\begin{tabular}{c|ccc|ccc}
\Xhline{1.0pt}

\multirow{2}{*}{class cost} & \multicolumn{3}{c|}{Novel}  & \multicolumn{3}{c}{All} \\
\cline{2-7} 
 & AP$^{\rm m}$ & AR$_{10}^{\rm m}$ & AR$_{100}^{\rm m}$
 & AP$^{\rm m}$ & AR$_{10}^{\rm m}$ & AR$_{100}^{\rm m}$ \\
\Xhline{1.0pt}

\xmark & 8.9 & 17.0 & 40.7 & 16.6 & 26.2 & 51.4 \\
\rowcolor{lightgray!38}\cmark & 12.8 & 19.4 & 40.6 & 28.0 & 32.4 & 51.5 \\

\Xhline{1.0pt}
\end{tabular}}}
    \label{tab:ab_class}
    \vspace{-0.5mm}
    \centering
    \captionsetup{width=0.95\linewidth}
    \captionof{table}{Ablation on the values of $k_{1}$ and $k_{2}$ in contrastive learning.}
    \vspace{1mm}
    \setlength{\tabcolsep}{1.5mm}{
    \scalebox{1.0}{
    \small
    \begin{tabular}{cc|ccc|ccc}
\Xhline{1.0pt}

\multirow{2}{*}{$k_{1}$} & \multirow{2}{*}{$k_{2}$} & \multicolumn{3}{c|}{Novel}  & \multicolumn{3}{c}{All} \\
\cline{3-8} 
 &
 & AP$^{\rm m}$ & AR$_{10}^{\rm m}$ & AR$_{100}^{\rm m}$ 
 & AP$^{\rm m}$ & AR$_{10}^{\rm m}$ & AR$_{100}^{\rm m}$ \\
\Xhline{1.0pt}

1 & 100 & 11.4 & 19.7 & 40.1 & 25.5 & 31.6 & 50.7 \\
\rowcolor{lightgray!38}10 & 100 & 12.8 & 19.4 & 40.6 & 28.0 & 32.4 & 51.5 \\ 
50 & 100 & 8.9 & 17.0 & 40.3 & 19.3 & 28.9 & 51.5 \\
10 & 20 & 12.1 & 19.2 & 40.3 & 27.4 & 32.2 & 51.5 \\
10 & 200 & 13.0 & 20.8 & 39.7 & 28.3 & 32.8 & 50.6 \\

\Xhline{1.0pt}
\end{tabular}}}
    \label{tab:ab_k1k2}
\end{minipage}
\vspace{-4mm}
\end{figure}

\myparagraph{Classification Cost for Sample Selection in Contrastive Learning.} To evaluate the effect of classification cost in the contrastive learning, we set $C_{cls}=0$ in Eq. (\ref{eq:matching_cost}) for the ablation. From Table~\ref{tab:ab_class}, we observe that performance drops drastically without classification cost. Such phenomenon stands with the view that classification score is crucial for two potential reasons. First, the classification cost ensures the network's consistency in assigning labels during both contrastive learning and network training. Second, the localization cost alone will introduce those predictions closest to the ground-truths as positive samples, while classification cost helps choose more discriminative samples.

\myparagraph{The Values of $k_{1}$ and $k_{2}$ in Contrastive Learning.} To study the impact on the number of positive and negative samples, we provide the ablation results of $k_{1}$ and $k_{2}$ in Table~\ref{tab:ab_k1k2}. The first three rows show that increasing the number of positive samples can result in a large number of false positives, which negatively affect the average precision (AP). By comparing line 2-4-5, it indicates that more negative samples benefit AP while hurt AR@100. This is reasonable because more negative samples can aid contrastive learning in generating more distinct representations; however, it may also incorrectly identify real objects as negative samples and suppress them as background.

\begin{figure}[t]
\vspace{-1mm}
\hspace{0mm}
\includegraphics[width=0.50\textwidth]{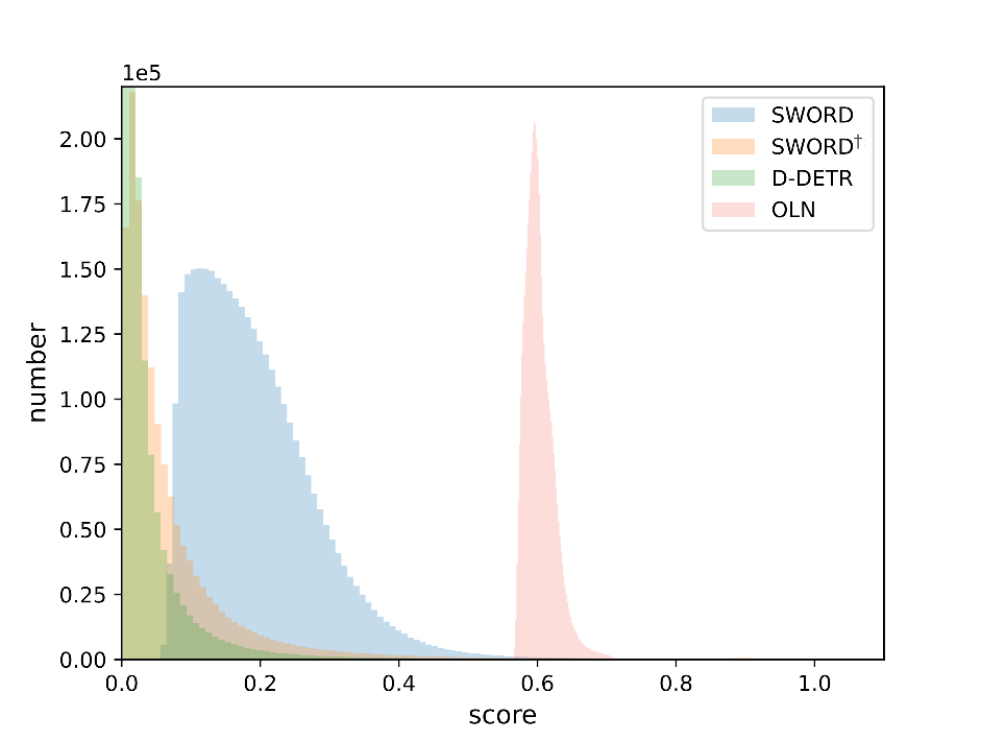}
\vspace{-5mm}
\caption{Comparing score distributions of proposals for different methods on COCO~\cite{lin2014coco} validation set. `D-DETR' represents Deformable-DETR~\cite{zhu2020deformable-detr}. All models are trained with 20 PASCAL-VOC classes. For fair comparisons, we select the top-100 proposals per image for all the methods. }
\label{fig:scores}
\vspace{-6mm}
\end{figure}

\vspace{-1mm}
\subsection{Visualization}
\vspace{-1mm}

In Figure~\ref{fig:scores}, we visualize the score distribution of different methods in VOC to non-VOC setting. Deformable-DETR~\cite{zhu2020deformable-detr} can only find out the seen category objects and thus its score distribution is mainly located on the low-scoring areas. OLN~\cite{kim2022oln} is trained with positive samples, making it merely produce the high-scoring proposals. Although it reveals the open-world ability to locate \textit{novel} objects, it can not effectively discriminate the objects and background. The proposed \modelname{} not only displays the favorable open-world generalization but also provide distinct confidence scores for objects and background.

\vspace{-1mm}
\section{Conclusion}
\vspace{-1mm}

In this work, we present a Transformer-based framework, \modelname, for open-world instance segmentation. Specifically, we introduce the \texttt{stop-grad} operation to prevent the feature degradation of \textit{novel} objects and propose a contrastive learning strategy to enlarge the discriminating representations between objects and background. We also develop an extension model, \modelname{}$^{\dag}$, by exploiting the pseudo labels of \modelname{}.
Extensive experiments demonstrate that the proposed models achieve state-of-the-art performance in various open-world generalization setups.

\section*{Acknowledgements}

This paper is partially supported by the National Key R\&D Program of China No.2022ZD0161000 and the General Research Fund of Hong Kong No.17200622. The paper is supported in part by the National Natural Science Foundation of China under grant No.62293540, 62293542, U1903215 and the Fundamental Research Funds for the Central Universities No.DUT22ZD210.

{\small
\bibliographystyle{ieee_fullname}
\bibliography{egbib}

\begin{thebibliography}{10}\itemsep=-1pt

\bibitem{arbelaez2006uch}
Pablo Arbelaez.
\newblock Boundary extraction in natural images using ultrametric contour maps.
\newblock In {\em 2006 Conference on Computer Vision and Pattern Recognition
  Workshop (CVPRW'06)}, pages 182--182. IEEE, 2006.

\bibitem{arbelaez2010owt}
Pablo Arbelaez, Michael Maire, Charless Fowlkes, and Jitendra Malik.
\newblock Contour detection and hierarchical image segmentation.
\newblock {\em IEEE transactions on pattern analysis and machine intelligence},
  33(5):898--916, 2010.

\bibitem{bai2022region}
Yutong Bai, Xinlei Chen, Alexander Kirillov, Alan Yuille, and Alexander~C Berg.
\newblock Point-level region contrast for object detection pre-training.
\newblock In {\em Proceedings of the IEEE/CVF Conference on Computer Vision and
  Pattern Recognition}, pages 16061--16070, 2022.

\bibitem{bendale2015towards}
Abhijit Bendale and Terrance Boult.
\newblock Towards open world recognition.
\newblock In {\em Proceedings of the IEEE conference on computer vision and
  pattern recognition}, pages 1893--1902, 2015.

\bibitem{carion2020detr}
Nicolas Carion, Francisco Massa, Gabriel Synnaeve, Nicolas Usunier, Alexander
  Kirillov, and Sergey Zagoruyko.
\newblock End-to-end object detection with transformers.
\newblock In {\em European conference on computer vision}, pages 213--229.
  Springer, 2020.

\bibitem{caron2021dino}
Mathilde Caron, Hugo Touvron, Ishan Misra, Herv{\'e} J{\'e}gou, Julien Mairal,
  Piotr Bojanowski, and Armand Joulin.
\newblock Emerging properties in self-supervised vision transformers.
\newblock In {\em Proceedings of the IEEE/CVF international conference on
  computer vision}, pages 9650--9660, 2021.

\bibitem{carreira2017k400}
Joao Carreira and Andrew Zisserman.
\newblock Quo vadis, action recognition? a new model and the kinetics dataset.
\newblock In {\em proceedings of the IEEE Conference on Computer Vision and
  Pattern Recognition}, pages 6299--6308, 2017.

\bibitem{cen2021deep}
Jun Cen, Peng Yun, Junhao Cai, Michael~Yu Wang, and Ming Liu.
\newblock Deep metric learning for open world semantic segmentation.
\newblock In {\em Proceedings of the IEEE/CVF International Conference on
  Computer Vision}, pages 15333--15342, 2021.

\bibitem{chen2021cyclemlp}
Shoufa Chen, Enze Xie, Chongjian Ge, Ding Liang, and Ping Luo.
\newblock Cyclemlp: A mlp-like architecture for dense prediction.
\newblock {\em arXiv preprint arXiv:2107.10224}, 2021.

\bibitem{chen2020simclr}
Ting Chen, Simon Kornblith, Mohammad Norouzi, and Geoffrey Hinton.
\newblock A simple framework for contrastive learning of visual
  representations.
\newblock In {\em International conference on machine learning}, pages
  1597--1607. PMLR, 2020.

\bibitem{chen2022cae}
Xiaokang Chen, Mingyu Ding, Xiaodi Wang, Ying Xin, Shentong Mo, Yunhao Wang,
  Shumin Han, Ping Luo, Gang Zeng, and Jingdong Wang.
\newblock Context autoencoder for self-supervised representation learning.
\newblock {\em arXiv preprint arXiv:2202.03026}, 2022.

\bibitem{chen2021simsiam}
Xinlei Chen and Kaiming He.
\newblock Exploring simple siamese representation learning.
\newblock In {\em Proceedings of the IEEE/CVF Conference on Computer Vision and
  Pattern Recognition}, pages 15750--15758, 2021.

\bibitem{chen2021mocov3}
Xinlei Chen*, Saining Xie*, and Kaiming He.
\newblock An empirical study of training self-supervised vision transformers.
\newblock {\em arXiv preprint arXiv:2104.02057}, 2021.

\bibitem{cheng2022mask2former}
Bowen Cheng, Ishan Misra, Alexander~G Schwing, Alexander Kirillov, and Rohit
  Girdhar.
\newblock Masked-attention mask transformer for universal image segmentation.
\newblock In {\em Proceedings of the IEEE/CVF Conference on Computer Vision and
  Pattern Recognition}, pages 1290--1299, 2022.

\bibitem{deng2009imagenet}
Jia Deng, Wei Dong, Richard Socher, Li-Jia Li, Kai Li, and Li Fei-Fei.
\newblock Imagenet: A large-scale hierarchical image database.
\newblock In {\em 2009 IEEE conference on computer vision and pattern
  recognition}, pages 248--255. Ieee, 2009.

\bibitem{devries2017cutout}
Terrance DeVries and Graham~W Taylor.
\newblock Improved regularization of convolutional neural networks with cutout.
\newblock {\em arXiv preprint arXiv:1708.04552}, 2017.

\bibitem{dosovitskiy2020vit}
Alexey Dosovitskiy, Lucas Beyer, Alexander Kolesnikov, Dirk Weissenborn,
  Xiaohua Zhai, Thomas Unterthiner, Mostafa Dehghani, Matthias Minderer, Georg
  Heigold, Sylvain Gelly, et~al.
\newblock An image is worth 16x16 words: Transformers for image recognition at
  scale.
\newblock {\em arXiv preprint arXiv:2010.11929}, 2020.

\bibitem{everingham2010pascal}
Mark Everingham, Luc Van~Gool, Christopher~KI Williams, John Winn, and Andrew
  Zisserman.
\newblock The pascal visual object classes (voc) challenge.
\newblock {\em International journal of computer vision}, 88(2):303--338, 2010.

\bibitem{ge2021ota}
Zheng Ge, Songtao Liu, Zeming Li, Osamu Yoshie, and Jian Sun.
\newblock Ota: Optimal transport assignment for object detection.
\newblock In {\em Proceedings of the IEEE/CVF Conference on Computer Vision and
  Pattern Recognition}, pages 303--312, 2021.

\bibitem{ge2021yolox}
Zheng Ge, Songtao Liu, Feng Wang, Zeming Li, and Jian Sun.
\newblock Yolox: Exceeding yolo series in 2021.
\newblock {\em arXiv preprint arXiv:2107.08430}, 2021.

\bibitem{grill2020byol}
Jean-Bastien Grill, Florian Strub, Florent Altch{\'e}, Corentin Tallec, Pierre
  Richemond, Elena Buchatskaya, Carl Doersch, Bernardo Avila~Pires, Zhaohan
  Guo, Mohammad Gheshlaghi~Azar, et~al.
\newblock Bootstrap your own latent-a new approach to self-supervised learning.
\newblock {\em Advances in neural information processing systems},
  33:21271--21284, 2020.

\bibitem{gupta2019lvis}
Agrim Gupta, Piotr Dollar, and Ross Girshick.
\newblock Lvis: A dataset for large vocabulary instance segmentation.
\newblock In {\em Proceedings of the IEEE/CVF conference on computer vision and
  pattern recognition}, pages 5356--5364, 2019.

\bibitem{han2019learning}
Kai Han, Andrea Vedaldi, and Andrew Zisserman.
\newblock Learning to discover novel visual categories via deep transfer
  clustering.
\newblock In {\em Proceedings of the IEEE/CVF International Conference on
  Computer Vision}, pages 8401--8409, 2019.

\bibitem{han2020coclr}
Tengda Han, Weidi Xie, and Andrew Zisserman.
\newblock Self-supervised co-training for video representation learning.
\newblock {\em Advances in Neural Information Processing Systems},
  33:5679--5690, 2020.

\bibitem{he2022mae}
Kaiming He, Xinlei Chen, Saining Xie, Yanghao Li, Piotr Doll{\'a}r, and Ross
  Girshick.
\newblock Masked autoencoders are scalable vision learners.
\newblock In {\em Proceedings of the IEEE/CVF conference on computer vision and
  pattern recognition}, pages 16000--16009, 2022.

\bibitem{he2020moco}
Kaiming He, Haoqi Fan, Yuxin Wu, Saining Xie, and Ross Girshick.
\newblock Momentum contrast for unsupervised visual representation learning.
\newblock In {\em Proceedings of the IEEE/CVF conference on computer vision and
  pattern recognition}, pages 9729--9738, 2020.

\bibitem{he2017maskrcnn}
Kaiming He, Georgia Gkioxari, Piotr Doll{\'a}r, and Ross Girshick.
\newblock Mask r-cnn.
\newblock In {\em Proceedings of the IEEE international conference on computer
  vision}, pages 2961--2969, 2017.

\bibitem{he2016resnet}
Kaiming He, Xiangyu Zhang, Shaoqing Ren, and Jian Sun.
\newblock Deep residual learning for image recognition.
\newblock In {\em Proceedings of the IEEE conference on computer vision and
  pattern recognition}, pages 770--778, 2016.

\bibitem{huang2022good}
Haiwen Huang, Andreas Geiger, and Dan Zhang.
\newblock Good: Exploring geometric cues for detecting objects in an open
  world.
\newblock {\em arXiv preprint arXiv:2212.11720}, 2022.

\bibitem{kalluri2023udos}
Tarun Kalluri, Weiyao Wang, Heng Wang, Manmohan Chandraker, Lorenzo Torresani,
  and Du Tran.
\newblock Open-world instance segmentation: Top-down learning with bottom-up
  supervision.
\newblock {\em arXiv preprint arXiv:2303.05503}, 2023.

\bibitem{kaul2022label}
Prannay Kaul, Weidi Xie, and Andrew Zisserman.
\newblock Label, verify, correct: A simple few shot object detection method.
\newblock In {\em Proceedings of the IEEE/CVF Conference on Computer Vision and
  Pattern Recognition}, pages 14237--14247, 2022.

\bibitem{khosla2020supcontrast}
Prannay Khosla, Piotr Teterwak, Chen Wang, Aaron Sarna, Yonglong Tian, Phillip
  Isola, Aaron Maschinot, Ce Liu, and Dilip Krishnan.
\newblock Supervised contrastive learning.
\newblock {\em Advances in Neural Information Processing Systems},
  33:18661--18673, 2020.

\bibitem{kim2022oln}
Dahun Kim, Tsung-Yi Lin, Anelia Angelova, In~So Kweon, and Weicheng Kuo.
\newblock Learning open-world object proposals without learning to classify.
\newblock {\em IEEE Robotics and Automation Letters}, 7(2):5453--5460, 2022.

\bibitem{kingma2014adam}
Diederik~P Kingma and Jimmy Ba.
\newblock Adam: A method for stochastic optimization.
\newblock {\em arXiv preprint arXiv:1412.6980}, 2014.

\bibitem{kwon2020backpropagated}
Gukyeong Kwon, Mohit Prabhushankar, Dogancan Temel, and Ghassan AlRegib.
\newblock Backpropagated gradient representations for anomaly detection.
\newblock In {\em Computer Vision--ECCV 2020: 16th European Conference,
  Glasgow, UK, August 23--28, 2020, Proceedings, Part XXI 16}, pages 206--226.
  Springer, 2020.

\bibitem{li2022dn-detr}
Feng Li, Hao Zhang, Shilong Liu, Jian Guo, Lionel~M Ni, and Lei Zhang.
\newblock Dn-detr: Accelerate detr training by introducing query denoising.
\newblock In {\em Proceedings of the IEEE/CVF Conference on Computer Vision and
  Pattern Recognition}, pages 13619--13627, 2022.

\bibitem{lin2022vldet}
Chuang Lin, Peize Sun, Yi Jiang, Ping Luo, Lizhen Qu, Gholamreza Haffari,
  Zehuan Yuan, and Jianfei Cai.
\newblock Learning object-language alignments for open-vocabulary object
  detection.
\newblock {\em arXiv preprint arXiv:2211.14843}, 2022.

\bibitem{lin2017focal}
Tsung-Yi Lin, Priya Goyal, Ross Girshick, Kaiming He, and Piotr Doll{\'a}r.
\newblock Focal loss for dense object detection.
\newblock In {\em Proceedings of the IEEE international conference on computer
  vision}, pages 2980--2988, 2017.

\bibitem{lin2014coco}
Tsung-Yi Lin, Michael Maire, Serge Belongie, James Hays, Pietro Perona, Deva
  Ramanan, Piotr Doll{\'a}r, and C~Lawrence Zitnick.
\newblock Microsoft coco: Common objects in context.
\newblock In {\em European conference on computer vision}, pages 740--755.
  Springer, 2014.

\bibitem{liu2022dab-detr}
Shilong Liu, Feng Li, Hao Zhang, Xiao Yang, Xianbiao Qi, Hang Su, Jun Zhu, and
  Lei Zhang.
\newblock Dab-detr: Dynamic anchor boxes are better queries for detr.
\newblock {\em arXiv preprint arXiv:2201.12329}, 2022.

\bibitem{liu2021ubteacher}
Yen-Cheng Liu, Chih-Yao Ma, Zijian He, Chia-Wen Kuo, Kan Chen, Peizhao Zhang,
  Bichen Wu, Zsolt Kira, and Peter Vajda.
\newblock Unbiased teacher for semi-supervised object detection.
\newblock {\em arXiv preprint arXiv:2102.09480}, 2021.

\bibitem{liu2021swin}
Ze Liu, Yutong Lin, Yue Cao, Han Hu, Yixuan Wei, Zheng Zhang, Stephen Lin, and
  Baining Guo.
\newblock Swin transformer: Hierarchical vision transformer using shifted
  windows.
\newblock In {\em Proceedings of the IEEE/CVF International Conference on
  Computer Vision}, pages 10012--10022, 2021.

\bibitem{milletari2016vnet}
Fausto Milletari, Nassir Navab, and Seyed-Ahmad Ahmadi.
\newblock V-net: Fully convolutional neural networks for volumetric medical
  image segmentation.
\newblock In {\em 2016 fourth international conference on 3D vision (3DV)},
  pages 565--571. IEEE, 2016.

\bibitem{peyre2019computational}
Gabriel Peyr{\'e}, Marco Cuturi, et~al.
\newblock Computational optimal transport: With applications to data science.
\newblock {\em Foundations and Trends{\textregistered} in Machine Learning},
  11(5-6):355--607, 2019.

\bibitem{qi2021open}
Lu Qi, Jason Kuen, Yi Wang, Jiuxiang Gu, Hengshuang Zhao, Zhe Lin, Philip Torr,
  and Jiaya Jia.
\newblock Open-world entity segmentation.
\newblock {\em arXiv preprint arXiv:2107.14228}, 2021.

\bibitem{qiao2021defrcn}
Limeng Qiao, Yuxuan Zhao, Zhiyuan Li, Xi Qiu, Jianan Wu, and Chi Zhang.
\newblock Defrcn: Decoupled faster r-cnn for few-shot object detection.
\newblock In {\em Proceedings of the IEEE/CVF International Conference on
  Computer Vision}, pages 8681--8690, 2021.

\bibitem{rezatofighi2019giou}
Hamid Rezatofighi, Nathan Tsoi, JunYoung Gwak, Amir Sadeghian, Ian Reid, and
  Silvio Savarese.
\newblock Generalized intersection over union: A metric and a loss for bounding
  box regression.
\newblock In {\em Proceedings of the IEEE/CVF conference on computer vision and
  pattern recognition}, pages 658--666, 2019.

\bibitem{saito2021ldet}
Kuniaki Saito, Ping Hu, Trevor Darrell, and Kate Saenko.
\newblock Learning to detect every thing in an open world.
\newblock {\em arXiv preprint arXiv:2112.01698}, 2021.

\bibitem{shao2019objects365}
Shuai Shao, Zeming Li, Tianyuan Zhang, Chao Peng, Gang Yu, Xiangyu Zhang, Jing
  Li, and Jian Sun.
\newblock Objects365: A large-scale, high-quality dataset for object detection.
\newblock In {\em Proceedings of the IEEE/CVF international conference on
  computer vision}, pages 8430--8439, 2019.

\bibitem{shi2000gpb}
Jianbo Shi and Jitendra Malik.
\newblock Normalized cuts and image segmentation.
\newblock {\em IEEE Transactions on pattern analysis and machine intelligence},
  22(8):888--905, 2000.

\bibitem{sohn2020stac}
Kihyuk Sohn, Zizhao Zhang, Chun-Liang Li, Han Zhang, Chen-Yu Lee, and Tomas
  Pfister.
\newblock A simple semi-supervised learning framework for object detection.
\newblock {\em arXiv preprint arXiv:2005.04757}, 2020.

\bibitem{sun2022gradient}
Jingbo Sun, Li Yang, Jiaxin Zhang, Frank Liu, Mahantesh Halappanavar, Deliang
  Fan, and Yu Cao.
\newblock Gradient-based novelty detection boosted by self-supervised binary
  classification.
\newblock In {\em Proceedings of the AAAI Conference on Artificial
  Intelligence}, volume~36, pages 8370--8377, 2022.

\bibitem{tian2023spark}
Keyu Tian, Yi Jiang, Qishuai Diao, Chen Lin, Liwei Wang, and Zehuan Yuan.
\newblock Designing bert for convolutional networks: Sparse and hierarchical
  masked modeling.
\newblock {\em arXiv preprint arXiv:2301.03580}, 2023.

\bibitem{tian2020condinst}
Zhi Tian, Chunhua Shen, and Hao Chen.
\newblock Conditional convolutions for instance segmentation.
\newblock In {\em European conference on computer vision}, pages 282--298.
  Springer, 2020.

\bibitem{tolstikhin2021mlp-mixer}
Ilya~O Tolstikhin, Neil Houlsby, Alexander Kolesnikov, Lucas Beyer, Xiaohua
  Zhai, Thomas Unterthiner, Jessica Yung, Andreas Steiner, Daniel Keysers,
  Jakob Uszkoreit, et~al.
\newblock Mlp-mixer: An all-mlp architecture for vision.
\newblock {\em Advances in Neural Information Processing Systems},
  34:24261--24272, 2021.

\bibitem{vaswani2017transformer}
Ashish Vaswani, Noam Shazeer, Niki Parmar, Jakob Uszkoreit, Llion Jones,
  Aidan~N Gomez, {\L}ukasz Kaiser, and Illia Polosukhin.
\newblock Attention is all you need.
\newblock {\em Advances in neural information processing systems}, 30, 2017.

\bibitem{villani2009optimal}
C{\'e}dric Villani.
\newblock {\em Optimal transport: old and new}, volume 338.
\newblock Springer, 2009.

\bibitem{wang2023openinst}
Cheng Wang, Guoli Wang, Qian Zhang, Peng Guo, Wenyu Liu, and Xinggang Wang.
\newblock Openinst: A simple query-based method for open-world instance
  segmentation.
\newblock {\em arXiv preprint arXiv:2303.15859}, 2023.

\bibitem{wang2022ggn}
Weiyao Wang, Matt Feiszli, Heng Wang, Jitendra Malik, and Du Tran.
\newblock Open-world instance segmentation: Exploiting pseudo ground truth from
  learned pairwise affinity.
\newblock In {\em Proceedings of the IEEE/CVF Conference on Computer Vision and
  Pattern Recognition}, pages 4422--4432, 2022.

\bibitem{wang2021uvo}
Weiyao Wang, Matt Feiszli, Heng Wang, and Du Tran.
\newblock Unidentified video objects: A benchmark for dense, open-world
  segmentation.
\newblock In {\em Proceedings of the IEEE/CVF International Conference on
  Computer Vision}, pages 10776--10785, 2021.

\bibitem{wang2021pvt}
Wenhai Wang, Enze Xie, Xiang Li, Deng-Ping Fan, Kaitao Song, Ding Liang, Tong
  Lu, Ping Luo, and Ling Shao.
\newblock Pyramid vision transformer: A versatile backbone for dense prediction
  without convolutions.
\newblock In {\em Proceedings of the IEEE/CVF International Conference on
  Computer Vision}, pages 568--578, 2021.

\bibitem{wang2021densecl}
Xinlong Wang, Rufeng Zhang, Chunhua Shen, Tao Kong, and Lei Li.
\newblock Dense contrastive learning for self-supervised visual pre-training.
\newblock In {\em Proceedings of the IEEE/CVF Conference on Computer Vision and
  Pattern Recognition}, pages 3024--3033, 2021.

\bibitem{wu2022idol}
Junfeng Wu, Qihao Liu, Yi Jiang, Song Bai, Alan Yuille, and Xiang Bai.
\newblock In defense of online models for video instance segmentation.
\newblock {\em arXiv preprint arXiv:2207.10661}, 2022.

\bibitem{xie2021detco}
Enze Xie, Jian Ding, Wenhai Wang, Xiaohang Zhan, Hang Xu, Peize Sun, Zhenguo
  Li, and Ping Luo.
\newblock Detco: Unsupervised contrastive learning for object detection.
\newblock In {\em Proceedings of the IEEE/CVF International Conference on
  Computer Vision}, pages 8392--8401, 2021.

\bibitem{xie2020self}
Qizhe Xie, Minh-Thang Luong, Eduard Hovy, and Quoc~V Le.
\newblock Self-training with noisy student improves imagenet classification.
\newblock In {\em Proceedings of the IEEE/CVF conference on computer vision and
  pattern recognition}, pages 10687--10698, 2020.

\bibitem{xu2021soft-teacher}
Mengde Xu, Zheng Zhang, Han Hu, Jianfeng Wang, Lijuan Wang, Fangyun Wei, Xiang
  Bai, and Zicheng Liu.
\newblock End-to-end semi-supervised object detection with soft teacher.
\newblock In {\em Proceedings of the IEEE/CVF International Conference on
  Computer Vision}, pages 3060--3069, 2021.

\bibitem{xue2022single}
Xizhe Xue, Dongdong Yu, Lingqiao Liu, Yu Liu, Satoshi Tsutsui, Ying Li, Zehuan
  Yuan, Ping Song, and Mike~Zheng Shou.
\newblock Single-stage open-world instance segmentation with cross-task
  consistency regularization.
\newblock 2022.

\bibitem{yan2023uninext}
Bin Yan, Yi Jiang, Jiannan Wu, Dong Wang, Ping Luo, Zehuan Yuan, and Huchuan
  Lu.
\newblock Universal instance perception as object discovery and retrieval.
\newblock In {\em Proceedings of the IEEE/CVF Conference on Computer Vision and
  Pattern Recognition}, pages 15325--15336, 2023.

\bibitem{yang2021objects}
Shuo Yang, Peize Sun, Yi Jiang, Xiaobo Xia, Ruiheng Zhang, Zehuan Yuan, Changhu
  Wang, Ping Luo, and Min Xu.
\newblock Objects in semantic topology.
\newblock {\em arXiv preprint arXiv:2110.02687}, 2021.

\bibitem{zhou2021ibot}
Jinghao Zhou, Chen Wei, Huiyu Wang, Wei Shen, Cihang Xie, Alan Yuille, and Tao
  Kong.
\newblock ibot: Image bert pre-training with online tokenizer.
\newblock {\em arXiv preprint arXiv:2111.07832}, 2021.

\bibitem{zhu2020deformable-detr}
Xizhou Zhu, Weijie Su, Lewei Lu, Bin Li, Xiaogang Wang, and Jifeng Dai.
\newblock Deformable detr: Deformable transformers for end-to-end object
  detection.
\newblock {\em arXiv preprint arXiv:2010.04159}, 2020.

\bibitem{zoph2020rethinking}
Barret Zoph, Golnaz Ghiasi, Tsung-Yi Lin, Yin Cui, Hanxiao Liu, Ekin~Dogus
  Cubuk, and Quoc Le.
\newblock Rethinking pre-training and self-training.
\newblock {\em Advances in neural information processing systems},
  33:3833--3845, 2020.

\end{thebibliography}
}

\clearpage
\begin{appendices}
\section{More Related Work} \label{sec:literature}

\myparagraph{Open-world Instance Segmentation.} This part supplements the related work in main paper. As is pointed out, closed-world models treat the un-annotated objects as background during training and thus can not discover the \textit{novel} objects from backdrop during inference. In order to solve the problem, there have emerged many advanced open-world works~\cite{kim2022oln, saito2021ldet, wang2022ggn, huang2022good, wang2023openinst, kalluri2023udos} recently.

OLN~\cite{kim2022oln} proposes to replace the classification head with localization quality head (\textit{e.g.}, IoU head) to predict the proposal scores. Because it is only trained with positive samples, OLN would not suppress \textit{novel} objects as background.
LDET~\cite{saito2021ldet} addresses the task from the perspective of synthesizing images without hidden objects as the training source. Specifically, LDET proposes a data augmentation named BackErase, which pastes the annotated objects on a background image sampled from a small region. In this way, objects and background can be clearly distinguished. 
GGN~\cite{wang2022ggn} proposes to solve the problem by exploiting the pseudo ground-truth of learned pairwise affinity. It first uses the classical grouping algorithms~\cite{arbelaez2006uch, arbelaez2010owt, shi2000gpb} to generate pseudo masks from pairwise affinity predictor. Then, Mask-RCNN~\cite{he2017maskrcnn} is trained with the augmented annotations. 
GOOD~\cite{huang2022good} exploits the geometric cues such as depth and normals, predicted by the monocular estimators, as the additional training sets. The authors train the OLN-like proposal network for pseudo-labeling novel objects from these training source, which shows significant effectiveness. UDOS~\cite{kalluri2023udos} combines classical bottom-up grouping with top-down learning framework. It utilizes the affinity-based grouping and refinement modules to gather the part-masks as the robust instance-level segmentations. OpenInst~\cite{wang2023openinst} is a concurrent work that uses the query-based detector for open-world instance segmentation.


\begin{algorithm}[t]
\caption{Pseudo-code of Contrastive Learning.}
\label{alg:code}
\algcomment{\fontsize{7.2pt}{0em}\selectfont \texttt{mm}: matrix multiplication; \texttt{cat}: concatenation.
}
\begin{lstlisting}[language=Python]
# transformer: the transformer network
# f_q, f_k: contrastive head for query and key
# queue: store the object embeddings, KxC
# m: momentum
# t: temperature

f_q.params = f_k.params  # initialize

# load an image and its targets
for image, targets in loader: 
    
    # get the query predictions 
    queries = transformer.forward(image)
    q = f_q.forward(queries) # NxC
    k = f_k.forward(queries) # NxC
    
    # for each ground-truth object
    for target in targets: 
        queue = queue.detach() 
        v = mean(queue, dim=0) # object center, 1xC
        
        # positive and negative selection, 
        # according to Eq.(1) in Appendix
        k_pos_id, k_neg_id = SimOTA(queries, target) 

        k_pos = k.index_select(k_pos_id) # k1xC
        k_neg = k.index_select(k_neg_id) # (k2-k1)xC
        
        # positive logits: 1xk1
        l_pos = mm(v, k_pos.transpose(0,1))
        
        # negative logits: 1x(k2-k1)
        l_neg = mm(v, k_neg.transpose(0,1))
        
        # logits: 1x[k1+(k2-l1)]
        logits = cat([l_pos, l_neg], dim=1)
        
        # contrastive loss, Eq.(2) in main paper
        labels = cat([ones(k1), zeros(k2-k1)], dim=0)
        loss = ContrastiveLoss(logits/t, labels)
        
        # Adam update: transformer and f_k
        loss.backward()
        update(transformer.params)
        update(f_k.params)
        
        # momentum update: f_q
        f_q.params = m*f_q.params+(1-m)*f_k.params
        
    # find the best matched queries for ground-truths
    query_ids = BipartiteMatch(queries, targets)
    
    # update queue
    q_c = q.index_select(query_id)
    enqueue(queue, q_c)
    dequeue(queue)
        
        
\end{lstlisting}
\end{algorithm}

\section{Architecture}\label{sec:architecture}

\begin{figure*}[t]
\vspace{-2mm}
\begin{center}
\includegraphics[width=0.92\textwidth]{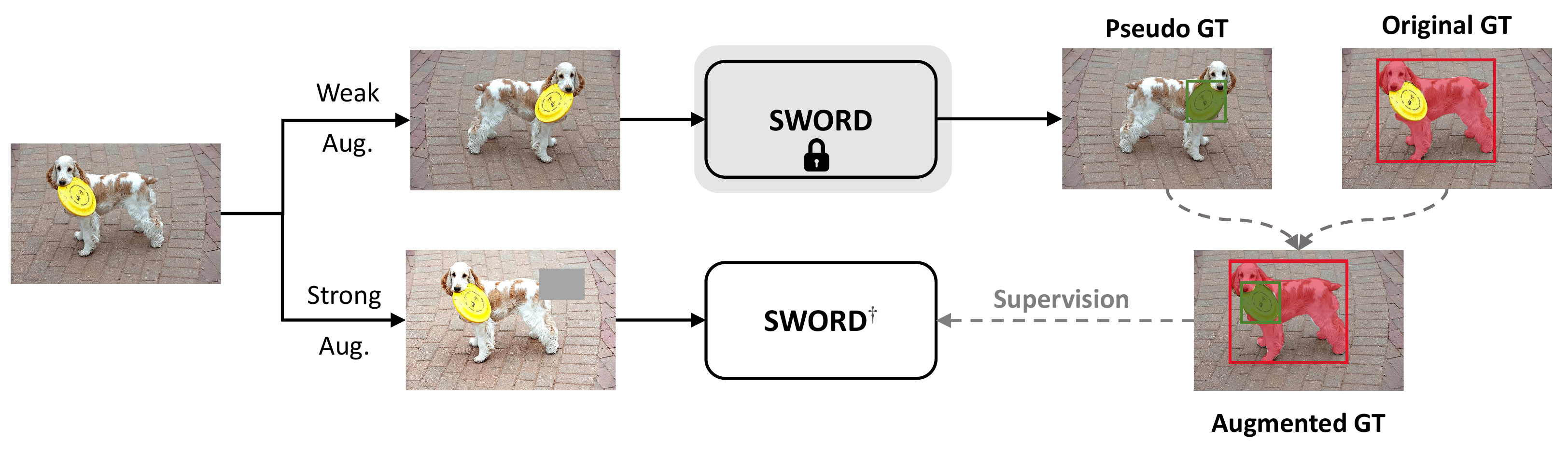}
\end{center}
\vspace{-5mm}
\caption{\textbf{The pipeline of pseudo ground-truth training}. The pretrained \modelname{} is first adopted to generate the pseudo boxes/masks. Then the top-scoring predictions are merged with the original annotations. Finally, \modelname{}$^{\dag}$ is trained under the supervision of augmented ground-truths. Note that \modelname{}$^{\dag}$ uses exactly the same architecture as Deformable-DETR.}
\label{fig:self-training}
\vspace{-4mm}
\end{figure*}

\subsection{Contrastive Learning}

We provide the pseudo-code of contrastive learning in Algorithm~\ref{alg:code}. The object center plays the role of query. Positive and negative samples are from the query embeddings for each image. The contrastive learning framework is only used for training and is simply abandoned during inference.

\myparagraph{Universal Object Queue.} The universal object queue $Q=\left [ q_{1}, q_{2}, ..., q_{K} \right ] \in \mathbb{R}^{K \times C}$ stores the object embeddings, where $K$ is the queue size and $C$ is the channel dimension of embeddings. The queue is randomly initialized. In each training iteration, the query embeddings of those predictions best matching the ground-truths are enqueue and the oldest ones are dequeue. Notably, these embeddings are computed by the slowly updated contrastive head $f_{q}$ to ensure the stability of universal object queue.

\myparagraph{Sample Selection.} For contrastive learning, we adopt the SimOTA~\cite{ge2021yolox, wu2022idol} strategy to dynamically select the positive and negative samples according to the matching cost. Given an image, we compute the matching cost between the $i$-th prediction $p_{i}$ and the $j$-th ground-truth $g_{j}$ as

\vspace{-2mm}
\begin{equation}
    \label{eq:cost}
    C^{ij} = \lambda_{cls} \cdot C^{ij}_{cls} + \lambda_{L1} C^{ij}_{L1} + \lambda_{giou} C^{ij}_{giou}
    \vspace{0mm}
\end{equation}

\noindent
where $\lambda_{cls}$, $\lambda_{L1}$ and $\lambda_{giou}$ are the coefficients. $C^{ij}_{cls}$ is Focal loss~\cite{lin2017focal}, and $C^{ij}_{box}$ is a combination of the $\mathcal{L}_{1}$ loss and generalized IoU loss~\cite{rezatofighi2019giou}. For the ground-truth $g_{j}$, we sum up the top 10 IoU values to get $k_{1}$ and the top 100 IoU values to get $k_{2}$. Then, we take the top $k_{1}$ predictions with the lowest cost as positive samples. To improve the embedding quality of negative samples, we choose the top $k_{2}$ predictions with the lowest cost and exclude the first $k_{1}$ ones. The left $k_{2}-k_{1}$ predictions are the hard negatives. We use the regularly updated contrastive head $f_{k}$ to compute their embeddings and form the positive set $\mathcal{K}^{+}$ and negative set $\mathcal{K}^{-}$.

\subsection{Pseudo Ground-truth Training}
\vspace{-1mm}

\myparagraph{Details.} The previous work GGN~\cite{wang2022ggn} shows that the pseudo labeling method can greatly boost the performance of Mask-RCNN in open world. Inspired by this work, we also develop an extension model, \modelname{}$^{\dag}$, by exploiting the pseudo ground-truth of \modelname{}. As shown in Figure~\ref{fig:self-training}, we first use \modelname{} to generate the pseudo boxes/masks. Then the top-scoring predictions are merged with the original annotations to form the augmented ground-truths, which plays the role of supervision to train the \modelname{}$^{\dag}$. Note that \modelname{}$^{\dag}$ uses exactly the same architecture as closed-world model Deformable-DETR~\cite{zhu2020deformable-detr}.

In the pseudo labeling process, we empirically find that using the IoU scores of \modelname{} leads to better learning results. And the merge process directly follows the existing practice~\cite{wang2022ggn}. Specifically, we first set the NMS value as 0.3 for \modelname{} to remove most predictions. Considering that the pseudo labels should focus on covering the \textit{novel} objects, we discard those proposals having the box IoU greater than 0.5 with the annotated objects. Finally, the top-$k$ predictions are kept as pseudo ground-truths.

\myparagraph{Data Augmentation.} Data augmentation has been demonstrated to play an important role in the self-training~\cite{xie2020self, kaul2022label, zoph2020rethinking} and semi-supervised methods~\cite{sohn2020stac, liu2021ubteacher, xu2021soft-teacher}. Following \cite{liu2021ubteacher}, we use the random horizontal flip for weak augmentation. And the strong augmentation includes random color jittering, grayscale, Gaussian blur and random cutout~\cite{devries2017cutout}.

\begin{table*}[t]
\centering
\renewcommand\arraystretch{1.10} 
\setlength{\tabcolsep}{6.5pt}    
\captionsetup{width=0.95\linewidth}
\caption{\textbf{Ablation on strong augmentation in pseudo ground-truth training}. We evaluate the models in COCO to UVO and VOC to non-VOC setups. And the results are reported on the \textit{novel} objects.}
\vspace{-2mm}




\begin{tabular}{c|ccc|ccc|ccc|ccc}
\Xhline{1.0pt}

\multirow{2}{*}{Strong Aug.} & \multicolumn{6}{c|}{\textbf{COCO to UVO}}  & \multicolumn{6}{c}{\textbf{VOC to non-VOC}} \\
\cline{2-13} 
 & AP$^{\rm b}$ & AR$_{10}^{\rm b}$ & AR$_{100}^{\rm b}$
 & AP$^{\rm m}$ & AR$_{10}^{\rm m}$ & AR$_{100}^{\rm m}$ 
 & AP$^{\rm b}$ & AR$_{10}^{\rm b}$ & AR$_{100}^{\rm b}$ 
 & AP$^{\rm m}$ & AR$_{10}^{\rm m}$ & AR$_{100}^{\rm m}$ \\
\Xhline{1.0pt}

\xmark & 16.0 & 22.3 & 49.5 & 12.1 & 20.5 & 42.3 & 5.6 & 21.4 & 38.8 & 5.2 & 19.7 & 33.8 \\
\rowcolor{lightgray!38}\cmark & 16.6 & 22.7 & 50.0 & 12.7 & 20.9 & 42.8 & 6.2 & 22.0 & 40.0 & 5.8 & 20.2 & 34.9 \\

\Xhline{1.0pt}
\end{tabular}
\label{tab:ab_aug} 
\vspace{-3mm}
\end{table*}

\begin{table}[t]
\centering
\renewcommand\arraystretch{1.10} 
\setlength{\tabcolsep}{5.0pt}    
\caption{\textbf{Ablation on the EMA rate}. The results are based on the COCO to UVO setup.}
\vspace{-2mm}




\begin{tabular}{c| ccc|ccc}
\Xhline{1.0pt}

\multirow{2}{*}{EMA} & \multicolumn{3}{c|}{Novel}  & \multicolumn{3}{c}{All} \\
\cline{2-7} 
 & AP$^{\rm m}$ & AR$_{10}^{\rm m}$ & AR$_{100}^{\rm m}$ 
 & AP$^{\rm m}$ & AR$_{10}^{\rm m}$ & AR$_{100}^{\rm m}$ \\
\Xhline{1.0pt}

0.5 & 8.9 & 16.3 & 27.8 & 16.9 & 24.4 & 35.8 \\
0.9 & 11.3 & 19.2 & 37.4 & 24.3 & 30.4 & 47.8 \\
0.99 & 11.2 & 19.0 & 38.5 & 25.3 & 30.6 & 48.9 \\
\rowcolor{lightgray!38}0.999 & 12.8 & 19.4 & 40.6 & 28.0 & 32.4 & 51.5 \\
0.9999 & 11.9 & 18.6 & 40.7 & 28.4 & 32.7 & 52.0 \\
            
\Xhline{1.0pt}
\end{tabular}
\label{tab:ab_ema} 
\vspace{0mm}
\end{table}

\begin{table}[t]
\centering
\renewcommand\arraystretch{1.10} 
\setlength{\tabcolsep}{5.2pt}    
\caption{\textbf{Ablation on the universal object queue size}. The results are based on the VOC(COCO) to UVO setup.}
\vspace{-2mm}
\begin{tabular}{c|ccc|ccc}
\Xhline{1.0pt}

\multirow{2}{*}{Size}  & \multicolumn{3}{c|}{Novel}  & \multicolumn{3}{c}{All} \\
\cline{2-7} 
 & AP$^{\rm m}$ & AR$_{10}^{\rm m}$ & AR$_{100}^{\rm m}$ 
 & AP$^{\rm m}$ & AR$_{10}^{\rm m}$ & AR$_{100}^{\rm m}$ \\
\Xhline{1.0pt}

256 & 4.9 & 12.4 & 31.4 & 17.5 & 23.8 & 42.1 \\
1024 & 5.3 & 13.2 & 32.9 & 18.7 & 24.9 & 44.0 \\
\rowcolor{lightgray!38}4096 & 6.1 & 13.3 & 34.9 & 19.6 & 25.3 & 45.2 \\
8192 & 5.5 & 12.6 & 33.9 & 19.2 & 24.9 & 44.8 \\

\Xhline{1.0pt}
\end{tabular}
\label{tab:ab_size} 
\vspace{-4mm}
\end{table}

\vspace{-1mm}
\section{Implementation Details} \label{sec:impl_details}

\myparagraph{Model Details.} The model configurations mostly follow Deformable-DETR~\cite{zhu2020deformable-detr}. The Transformer has six encoders and six decoders with the hidden dimension of 256. To ensure a high recall, the object query number of \modelname{} is set to 2000 when trained on VOC classes and 1000 for all other settings. For contrastive learning, the size of universal object queue is set as 4096 and the exponential moving average (EMA) rate of the momentum contrastive head is 0.999. In the pseudo ground-truth training, \modelname{}$^{\dag}$ uses 1000 object queries for all the settings. ResNet-50~\cite{he2016resnet} is adopted as the backbone otherwise specified. 

\myparagraph{Training Details.} We use the Adam~\cite{kingma2014adam} optimizer with a base learning rate of $2 \times 10^{-4}$ and weight decay of $1 \times 10^{-4}$ for model training. All the models are trained on 8 GPUs with a batch size of 16. We present two models in this work, \modelname{} and \modelname{}$^{\dag}$. \modelname{} is trained for 80k iterations, with the learning rate decaying at the 60k-th iteration. As the VOC classes are partially annotated in COCO dataset, the model tends to overfit to the base classes. So we train \modelname{} from scratch when the training source is VOC. In all other settings, the backbone is initialized with the ImageNet~\cite{deng2009imagenet} pretrained weights. For \modelname{}$^{\dag}$, backbones always use the ImageNet pretrained weights for intialization. It undergoes 90k iterations of training, with the learning rate reduced by a factor of 10 at the 60k-th and 80k-th iterations. During training, we resize the input images such that the shortest side is at least 480 and at most 800, while the longest side is at most 1333. The loss coefficients are set as $\lambda_{cls}=2.0$, $\lambda_{cls}=2.0$, $\lambda_{L1}=5.0$, $\lambda_{mask}=2.0$, $\lambda_{dice}=5.0$ and $\lambda_{iou}=1.0$, respectively. All the models use the NMS value of 0.7 during inferene.

\vspace{-1mm}
\section{Additional Experimental Results} \label{sec:add_exp}

We provide additional experimental results to study the critical parameters for our method. The ablation studies are based on the COCO (80 classes) to UVO setup by default.

\subsection{Ablation on Contrastive Learning}
\vspace{-1mm}

\myparagraph{The Effect of EMA Rate.} The momentum update of the contrastive head can improve the consistency of the universal object queue. And a larger EMA rate allows the slower feature change. In Table~\ref{tab:ab_ema}, we present the experimental results with various EMA rate $\alpha$ from 0.5 to 0.9999. As illustrated in the first row, with the EMA rate of 0.5, the model gets relatively low results in both AP and AR metrics. This indicates that the model suffers from the detrimental effect of quick transformation of the object center. And the performance is greatly boosted with the EMA rate increases, \textit{e.g.}, the AP$^{\rm b}$ on \textit{all} objects achieves 6.9\% gain by increasing $\alpha$ from 0.5 to 0.9. We observe that the performance becomes stable when a larger EMA rate (\textit{e.g.}, $\alpha=0.999$) is applied.

\myparagraph{The Effect of Universal Object Queue Size.} In this study, we investigate the impact of the universal object queue size on the VOC(COCO) to UVO setup. Our findings are presented in Table~\ref{tab:ab_size}. We observe that when the queue size is increased from 256 to 4096, the model achieves a performance gain of 1.2 AP$^{\rm m}$ and 2.1 AP$^{\rm m}$ for novel and all objects, respectively. This improvement in performance may be attributed to the increased stability of the object center, which ensures that the object center captures the common characteristic of objects. However, we observe a decline in performance with further increases in the queue size, possibly due to the adverse effects of older object features on contrastive learning.

\subsection{Ablation on Pseudo Ground-truth Training}
\vspace{-1mm}

\begin{figure}[t]
\vspace{-4mm}
\begin{center}
\includegraphics[width=0.48\textwidth]{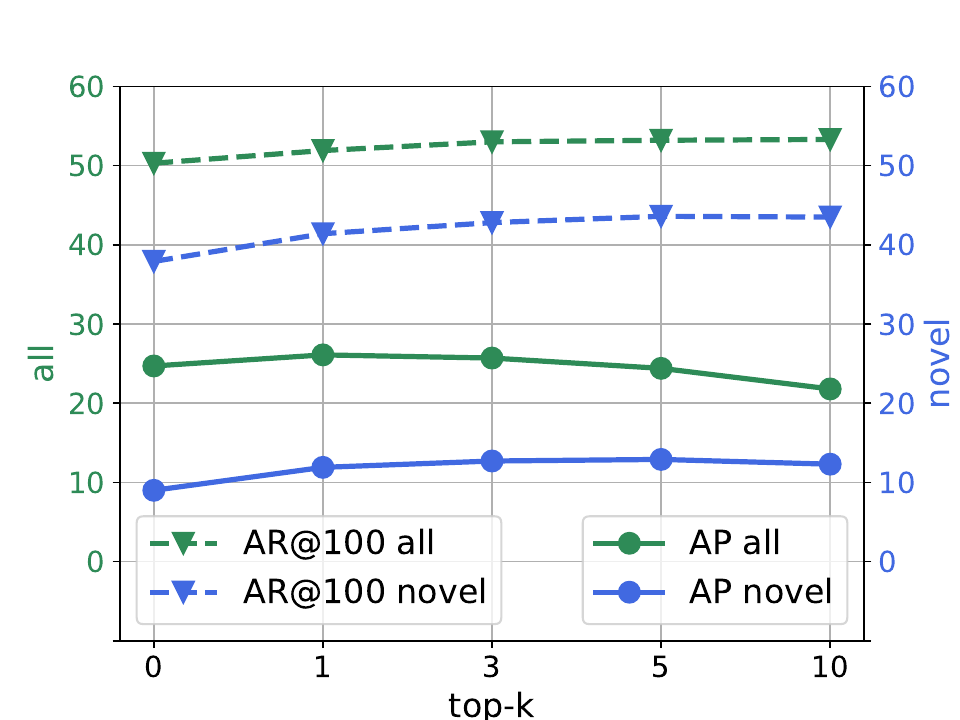}
\end{center}
\vspace{-4mm}
\caption{\textbf{The effect of top-k in pseudo ground-truth training}. The results are based on mask metrics in COCO to UVO setup.}
\label{fig:ab_topk}
\vspace{-5mm}
\end{figure}

\myparagraph{The Effect of Strong Augmentation.} To validate the effectiveness of strong augmentation in pseudo ground-truth training, we ablate the experiments in COCO to UVO and VOC to non-VOC settings, respectively. By comparing the two rows in Table~\ref{tab:ab_aug}, it is observed that the model could obtain better performance with the help of strong augmentation. Besides, we observe that the benefit of strong augmentation is more clear in VOC to non-VOC setup than COCO to UVO setup. The reason may attribute to the fact that the annotation density and class number of PASCAL-VOC are more limited, which requires the strong augmentation to generate more diverse training samples.

\myparagraph{The Effect of Pseudo Ground-truth Number.} The usage of pseudo ground-truth helps the closed-world models discover the \textit{novel} objects. However, it also introduces noisy supervision signals. To study the relationship between the model behavior and the number of pseudo ground-truth, we vary the number of $k$ for selecting the top-scoring predictions and plot the results in Figure~\ref{fig:ab_topk}. Here, we have 
the critical finding: 
\emph{More pseudo ground-truths benefit AR while hurting AP}. It can be seen that ARs keep improving with the increase of $k$, while AP for \textit{all} objects consistently degrades. AP for \textit{novel} objects also starts decreasing when $k$ reaches a large value (\textit{e.g.}, $k=10$). This is reasonable because more pseudo ground-truths will induce many false positive predictions. The results suggest that the value of top-$k$ should be carefully chosen to achieve the optimal balance between APs and ARs.

\subsection{More Ablation Studies}

\begin{table}[t]
\centering
\renewcommand\arraystretch{1.10} 
\setlength{\tabcolsep}{4.0pt}    
\caption{\textbf{Ablation on the query number.} The results are based on COCO to UVO setup. Our default settings are marked in \graybox{gray}.}
\begin{tabular}{c| ccc|ccc}
\Xhline{1.0pt}

\multirow{2}{*}{Query} & \multicolumn{3}{c|}{Novel}  & \multicolumn{3}{c}{All} \\
\cline{2-7} 
 & AP$^{\rm m}$ & AR$_{10}^{\rm m}$ & AR$_{100}^{\rm m}$ 
 & AP$^{\rm m}$ & AR$_{10}^{\rm m}$ & AR$_{100}^{\rm m}$ \\

\tabbl
\multicolumn{7}{l}{\small{\textbf{Deformable-DETR}}}  \\

300 & 8.9 & 16.1 & 37.1 & 24.4 & 29.8 & 49.7 \\
\rowcolor{lightgray!38}1000 & 9.0 & 16.7 & 37.9 & 24.7 & 30.1 & 50.3 \\
2000 & 8.6 & 15.8 & 37.9 & 24.7 & 30.0 & 50.3 \\

\tabbl
\multicolumn{7}{l}{\small{\textbf{SWORD}}}   \\

300 & 11.2 & 18.6 & 34.4 & 27.4 & 32.4 & 46.3 \\
\rowcolor{lightgray!38}1000 & 12.8 & 19.4 & 40.6 & 28.0 & 32.4 & 51.5 \\
2000 & 12.7 & 19.7 & 42.7 & 28.3 & 32.8 & 53.0 \\
            
\Xhline{1.0pt}
\end{tabular}
\label{tab:ab_query} 
\vspace{0mm}
\end{table}

\begin{table}[t]
\centering
\renewcommand\arraystretch{1.10} 
\setlength{\tabcolsep}{3.0pt}    
\caption{\textbf{Ablation on the backbones.} The results are based on COCO to UVO setup.}
\vspace{-2mm}





\begin{tabular}{c|ccc|ccc}
\Xhline{1.0pt}

\multirow{2}{*}{Backbone} & \multicolumn{3}{c|}{Novel}  & \multicolumn{3}{c}{All} \\
\cline{2-7} 
 & AP$^{\rm m}$ & AR$_{10}^{\rm m}$ & AR$_{100}^{\rm m}$
 & AP$^{\rm m}$ & AR$_{10}^{\rm m}$ & AR$_{100}^{\rm m}$ \\
\Xhline{1.0pt}

R50 & 12.8 & 19.5 & 40.6 & 28.0 & 32.4 & 51.5 \\
R101 & 12.6 & 19.9 & 41.3 & 29.5 & 33.4 & 52.7 \\
\hline

Swin-T & 12.2 & 19.5 & 40.8 & 29.4 & 33.4 & 52.0 \\
Swin-L & 13.5 & 20.5 & 41.2 & 34.3 & 37.0 & 54.1 \\

\Xhline{1.0pt}
\end{tabular}
\label{tab:ab_backbone} 
\vspace{-4mm}
\end{table}

\myparagraph{The Effect of Query Number.} We study the effect of query number for both Deformable-DETR and proposed \modelname{} in Table~\ref{tab:ab_query}. The results show that Deformable-DETR achieves a slight improvement in performance when the object query number is increased from 300 to 1000. However, the performance saturates at a query number of 1000, indicating that 1000 queries represent the upper limit for closed-world models to locate all objects in this open-world setup. In contrast, our proposed \modelname{} consistently achieves higher average recalls (ARs) as the query number increases. This performance profits can be attributed to the \texttt{stop-grad} operation, which prevents the suppression of novel objects and enables the network to discover them more effectively. It is worth noting that we use the same query number for both Deformable-DETR and \modelname{} in all experiments for fair comparisons.

\myparagraph{Do Stronger Backbones Benefit in Open-world?} There exists the consensus that stronger backbones~\cite{he2016resnet, dosovitskiy2020vit, wang2021pvt, liu2021swin, tolstikhin2021mlp-mixer, chen2021cyclemlp} could greatly increase the performance under the fully-supervised setup. Of particular interest, we examine with ResNet~\cite{he2016resnet} and Swin-Transformer~\cite{liu2021swin} to study the effect of using strong backbones in open-world scenario. Table~\ref{tab:ab_backbone} illustrates that model consistently performs better with increasing the size of backbones. Interestingly, we also observe that out-of-domain objects gets less benefit from stronger backbone than in-domain objects in the open-world. For example, by switching the backbone from Swin-Tiny to Swin-Large, the model enjoys the significant 4.9\% AP$^{\rm m}$ gain for \textit{all} objects while the advance is marginal for \textit{novel} objects (+1.3\% AP$^{\rm m}$).

\begin{table}[t]
\centering
\renewcommand\arraystretch{1.0} 
\setlength{\tabcolsep}{3.5pt}    
\small
\caption{\textbf{Ablation on the pseudo-label training for different models.} `w/ PL' represents the model is trained with the pseudo labels generated from the proposed SWORD. `D-DETR' denotes Deformable-DETR.}
\vspace{-5pt}

\begin{tabular}{cccccccc}
    \Xhline{1.0pt}
    \multirow{2}{*}{Method} & \multirow{2}{*}{w/ PL} & 
    \multicolumn{3}{c}{VOC to non-VOC}  & 
    \multicolumn{3}{c}{COCO to UVO} \\
    \cmidrule(lr){3-5} \cmidrule(lr){6-8}
     & & AP$^{\rm m}$ & AR$_{10}^{\rm m}$ & AR$_{100}^{\rm m}$ &
     AP$^{\rm m}$ & AR$_{10}^{\rm m}$ & AR$_{100}^{\rm m}$ \\
    \hline
    D-DETR & - & 2.2 & 10.2 & 22.7 & 9.0 & 16.7 & 37.4 \\
    D-DETR & \cmark & 5.8 & 20.2 & 34.9 & 12.7 & 20.9 & 42.8 \\
    \hline 
    SWORD & - & 4.8 & 15.7 & 30.2 & 12.8 & 19.4 & 40.6 \\
    SWORD  & \cmark & 5.9 & 20.9 & 36.2 & 13.3 & 21.4 & 43.5 \\
    \Xhline{1.0pt}
\end{tabular}

\label{tab:ab_self-train} 
\vspace{-8pt}
\end{table}

\myparagraph{Ablation on the Pseudo Ground-truth Training for Different Models.} We conduct the experiments using pseudo labels to train the proposed SWORD and display the results in Table~\ref{tab:ab_self-train}. We report the results on novel objects for both cross-category (VOC to non-VOC) and cross-dataset (COCO to UVO) generalizations. It is observed that the inclusion of pseudo-label training can further enhance the performance of SWORD, which also surpasses the results by using the standard Deformable-DETR for pseudo-label training. This highlights the strong ability of SWORD in discovering novel objects in the open-world scenario, proving the necessity of our designs.

\section{Visualization} \label{sec:visualization}

\begin{figure*}[h]
\begin{center}
\includegraphics[width=0.98\linewidth]{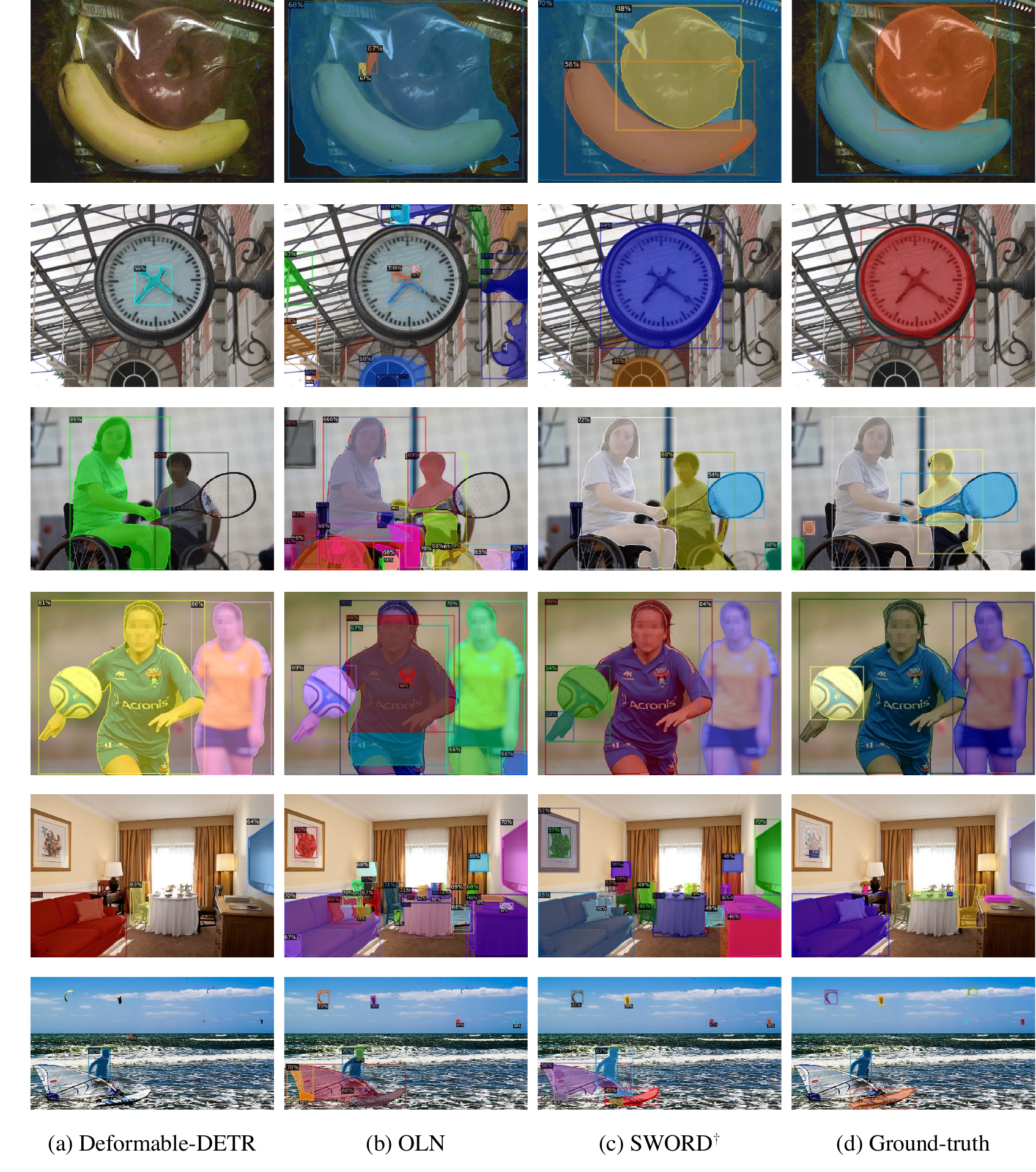}
\end{center}
\vspace{-2mm}
\caption{\textbf{Visualization examples in VOC to non-VOC setting}. All the models are trained on the 20 PASCAL-VOC classes of COCO dataset. The score thresholds for visualization are set as 0.45, 0.65 and 0.45 for Deformable-DETR~\cite{zhu2020deformable-detr}, OLN~\cite{kim2022oln} and SWORD$^{\dag}$, respectively. It is observed that Deformable-DETR is unable to segment the \textit{novel} objects and OLN produces many false positive predictions. Our model obviously provides the accurate and exhaustive segmentation masks.} 
\label{fig:vis_voc}
\vspace{0mm}
\end{figure*}

We visualize more examples in Figure~\ref{fig:vis_voc}. We demonstrate the superiority of proposed model in diverse scenes.

\end{appendices}

\end{document}